\documentclass[authoryear,5p]{elsarticle}

\usepackage{amsmath}
\usepackage{amssymb}
\usepackage{hyperref}

\makeatletter
\def\ps@pprintTitle{%
	\let\@oddhead\@empty
	\let\@evenhead\@empty
	\def\@oddfoot{\reset@font Preprint, not peer-reviewed\hfil\thepage\hfil}
	\let\@evenfoot\@oddfoot
}
\makeatother

\begin{document}

\begin{frontmatter}

\title{The Artificial Neural Twin - Process Optimization and Continual Learning in Distributed Process Chains} 

\author[IIS]{Johannes Emmert}\ead{johannes.emmert@iis.fraunhofer.de}     
\author[IIS]{Ronald Mendez}\ead{ronald.mendez@iis.fraunhofer.de}                
\author[IIS]{Houman Mirzaalian Dastjerdi}\ead{hooman.mirzaalian@gmail.com}   
\author[IIS,FAU]{Christopher Syben}\ead{christopher.syben@fau.de}   
\author[IIS,FAU]{Andreas Maier}\ead{andreas.maier@iis.fraunhofer.de}  

\address[IIS]{Fraunhofer IIS, Fraunhofer Institute for Integrated Circuits IIS, Division Development Center X-Ray Technology, Flugplatzstr. 75, 90768 F\"urth, Germany}                                           

\address[FAU]{Pattern Recognition Lab, Friedich-Alexander Universit\"at Erlangen-N\"urnberg, 91058 Erlangen, Germany}        

\begin{keyword}                           
Data-fusion; Continual Learning; Model predictive control;     Internet of things;     Multi sensor systems;  Distributed Learning;	Process Optimization;  	Decentralized and distributed control  
\end{keyword}                             

\begin{abstract}                          
Industrial process optimization and control is crucial to increase economic and ecologic efficiency. However, data sovereignty, differing goals, or the required expert knowledge for implementation impede holistic implementation. Further, the increasing use of data-driven AI-methods in process models and industrial sensory often requires regular fine-tuning to accommodate distribution drifts. We propose the Artificial Neural Twin, which combines concepts from model predictive control, deep learning, and sensor networks to address these issues. Our approach introduces differentiable data fusion to estimate the state of distributed process steps and their dependence on input data. By treating the interconnected process steps as a quasi neural-network, we can backpropagate loss gradients for process optimization or model fine-tuning to process parameters or AI models respectively. The concept is demonstrated on a virtual machine park simulated in Unity, consisting of bulk material processes in plastic recycling.
\end{abstract}

\end{frontmatter}

\section{Introduction}
Industrial process optimization is an important tool to increase economic and ecologic efficiency. However, implementation often requires experts to design a problem specific solution, and stakeholders (e.g. manufacturers of process machines) often hesitate to share interfaces or process models in the fear of losing data sovereignty. Additionally, process optimization requires knowledge about the states of the process steps to be optimized, amounting to a process monitoring task, that shares similar issues in terms of data sovereignty. At the same time AI-models become more and more abundant as process models in optimization and monitoring, or for data reduction in AI-sensors (e.g. camera sensors). External influences on the processed material stream, changes to the process or degradation of machinery may lead to decreased performance of such AI-models, requiring regular retraining on (expensive) updated datasets. Alternatively, continual learning \citep{Parisi2019} can be applied to constantly adapt the models to their changing environment by utilizing training stimuli from e.g. process monitoring or quality control. However, similar to the implementation of process optimization significant expert knowledge is required to avoid catastrophic forgetting in a specific implementation. This restriction can be relaxed by a common process interface that allows the experts (e.g. process machine manufacturers) to design process optimization and continuous learning routines specific to their machine but detached from the application environment. Any trained operator can then bind this machine into his process chain and set up process optimization and continual learning without risking process-safety or catastrophic forgetting. Further, the data exchange between distributed process steps is limited to information on the exchanged material streams, without sharing any model or structural data. To achieve this we combine concepts from  Model Predictive Control (MPC), supervised training methods, and decentral sensor networks into the Artificial Neural Twin (ANT). Process monitoring is implemented by decentral data fusion, which we implement in a differentiable manner. The differentiable data fusion allows for backpropagating loss gradients in analogy to neural network training through the physically distributed process steps, and thus provide training stimuli at the local nodes for process optimization or continual learning. An overview of the developments in the underlying topics is provided in the following.\par

MPC \citep{Kouvaritakis2015,2000} is developed and applied since decades to solve process optimization tasks. Put simply, MPC performs Kalman-filter-like predictions of future process states in a process chain (finite horizon) from a previous state, a process model, sensor data and postulated future settings of the controllable input parameters to the process. These predictions are then optimized in the controllable input parameters to reach the desired future state. After each time step the optimization is restarted, incorporating freshly acquired measurement data. Different variations of the method exist to account for nonlinear models, and different kinds of model uncertainties. For a more detailed overview please refer to \citep{Kouvaritakis2015,Christofides2013}. MPC is often applied to centrally control dynamic processes inside a company, meaning all data and models are available to a central node for optimization. If the process chain is distributed over companies or some proprietary process models belonging to third party machine manufacturers are used, central optimization and thus information disclosure is not desired \citep{schmitt2020architektur}. This limitation is resolved in decentralized and distributed MPC which divide the optimization problem into separate control loops with or without communication between the loops \citep{Christofides2013}. Nonetheless, the design of local cost functions or the inter-control-loop communication still requires coordination in order to achieve robust system operation.\par 
Instead of splitting the large central optimization problem into tasks for several small local control loops, it is also possible to solve the global MPC optimization problem in a distributed fashion. Scheu and Marquardt \citep{Scheu2011} propose coordinated distributed optimization by including linear sensitivity terms of all other control loops in each local control loop optimization, and thereby iteratively solving nonlinear global optimization problems. \par 
If gradients of the global cost function with respect to the local parameters are available, gradient descent methods can be applied directly to solve the global optimization problem. Many recent publications focus on distributed gradient descent methods in the context of federated learning \citep{Kairouz2021,Koloskova2021,Konecny2016,McMahan2016,Xin2020}, where models are available on every node, but the (training) data is only locally available. Related but more relevant in the context of MPC is the case of a distributed model to be trained/optimized. Examples of this setup are provided by Wang et al. for optimizing a communication relay network \citep{Wang2021a,Wang2021}. They regard the physical communication network as a quasi-neural network, with each relay node corresponding to a neuron, and the controllable physical parameters to be optimized corresponding to the weights. Backpropagation, typically known from deep-learning tasks \citep{Mitchell1997,maier2019gentle}, is then applied on the relay nodes to iteratively calculate gradient descent updates on the local weights. The concept of the quasi-neural network and backpropagation therethrough can be directly transferred to other physical process chains. Each process typically controlled by MPC is equvalent to a neuron or even a neural layer, the loss is given by the MPC parameter optimization, and the processes exchange loss gradients with their neighbors. Backpropagation takes over the role of inter control loop communication in distributed MPC, respectively the exchange of sensitivity data in the method of Scheu and Marquardt \citep{Scheu2011}. This direct analogy breaks when regarding the availability of state information at each process node: in digital systems the state of each process/relay/node is already known. In industrial processes the physical state of a process (material composition/quality, inner state of the machine etc.) typically first needs to be inferred. MPC solves this by incorporating current measurement data as a feedback into the Kalman prediction \citep{Kouvaritakis2015}. In a distributed setting, however, process nodes often require additional information from their neighbors to recover reliable state estimates. \par 

In the field of decentral sensor networks, this is implemented via distributed data fusion methods \citep{MartinLiggins2017,Qipeng2015,OlfatiSaber2007}. All nodes exchange messages with their neighbors communicating their state of knowledge on either the total or the relevant part of their state vector. Each node then fuses its own knowledge from previous predictions (i.e. Kalman filtering), sensor readings and process models with their neighbors knowledge. The consensus algorithms describing this fusion typically rely on statistical inference methods like Bayesian updating or extended Kalman filtering \citep{OlfatiSaber2007,Sommer2009,Qipeng2015}. \par

The ANT joins  the concepts of MPC, quasi-neural networks, and decentral data fusion. Each process step in the industrial chain represents an ANT-node, respectively neural layer, with the weights being the process parameters. For each time step a node estimates its current state vector and a prediction for the state in the next time step in a generalized nonlinear Kalman filter, formulated as statistical inference. The statistical inference incorporates nonlinear process and prediction models, sensor readings, controllable parameters, and information received from other nodes dynamically when available. To ensure numerical tractability of the problem, multivariate normal distributions are used as a probability model, thus also defining the exchanged information format as a mean (state) vector and the corresponding covariance matrix. Similar to MPC the state inference is solved sequentially from the soonest updated time step in the past to the last time step in the future, each yielding a maximum a posteriori (MAP) estimate and a corresponding posterior covariance matrix. The newly inferred information on the states are communicated to the neighbors. 
Through continuous communication between nodes a consensus on the material stream manifests and the actual optimization of the MPC problem can start.  \par 

The differentiable cost function to be optimized is dependent on the MAP estimate of the nodes' states and is calculated on a loss node. 
The loss node calculates the gradients of the costfunction with respect to each nodes state, and, if nonzero, sends the gradient to the corresponding node. To implement the chain rule of backpropagation on the receiving node, the Jacobian of state MAP estimate with respect to the fused mean vectors is required. The MAP estimate is defined implicitly by the data fusion problem. Hence, we introduce the concept of differentiable data fusion by applying the implicit function theorem to data fusion. The resultant Jacobian backpropagates the loss gradient, to the controllable parameters, previous time steps, and neighbors. After gradient messages from neighbors have decayed, a gradient descent step is calculated and the process parameters are updated accordingly. Then the process starts anew by communicating and thus finding a consensus on state vectors, iteratively reaching the optimum. Put differently, the information exchange and data fusion between physical process nodes amounts to the forward pass/inference in a neural network, while the differentiability of the data fusion allows for the backward pass in order to update weights.\par 
By utilizing appropriate loss sources (apart from MPC) like the data fusion residual or quality control spot checks, this general approach allows for backpropagating to sensors and process models, and thus continuously provides training stimuli for data driven process models to adapt to a changing environment. In order to keep the scope of this work tractable, we only outline the use for continual learning and list corresponding challenges, while the concept is demonstrated on a process optimization example.\par 
In the following sections we provide a more detailed overview of the different concepts and the implemented methods. Subsequently we demonstrate the method on a process optimization task in a virtual process chain of a plastic recycling facility. 
\section{Methods}

\label{sec:methods}

As a foundation for process optimization the state of all processes needs to be inferred by decentral data fusion between nodes. The Jacobians of the inferred states are then used to backpropagate loss gradients and iteratively optimize process parameters. The following sections first introduce underlying methods in a suitable form. Then the concept of differentiable data fusion is derived and embedded into the ANT algorithms.\par 
To avoid misunderstandings we want to clarify our nomenclature beforehand. By ANT node we refer to a software instance, that is able to read  sensor data and/or set process parameters on the associated process step. The ANT or ANT network refers to the network of ANT nodes that communicate to infer their states and optimize their models and process parameters.

\subsection{Backpropagation}
The supervised training of a neural network is the optimization of the trainiable weights of the neurons to minimize a loss function, that is dependent on the output of the neural network.
\begin{figure}
	\begin{center}
		\includegraphics[width=8.4cm]{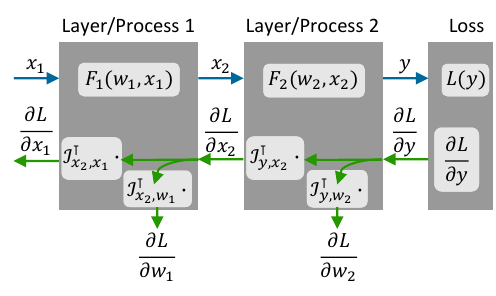}
		\caption{Backpropagation through a two layer network or two process steps. The forward pass (blue) amounts to the decentral data fusion and state estimation. The backward pass utilizes }
		\label{fig:backpropagation}                               
	\end{center}                                 
\end{figure}
Due to the large scale of typical deep-learning models the only feasible optimization method is gradient descent (in different variants), meaning the optimum is iteratively approached by stepping into the direction of the negative gradient. Backpropagation \citep{Mitchell1997,maier2019gentle} is a memory-wise and computationally efficient method to calculate the gradients of the loss with respect to the trainable weights. It relies on the repeated application of the chain rule, and reuses intermediate results, thereby avoiding to keep all Jacobians in memory. The procedure is illustrated in Figure \ref{fig:backpropagation} on a simplified neural network with two layers. 
We summarize the behavior of the layers as vector field functions $\mathbf{F}_1(\mathbf{w}_1,\mathbf{x}_1): \mathbb{R}^{N_{w1}}\times\mathbb{R}^{N_{x1}}\rightarrow\mathbb{R}^{N_{x2}}$ and $\mathbf{F}_2(\mathbf{w}_2,\mathbf{x}_2): \mathbb{R}^{N_{w2}}\times \mathbb{R}^{N_{x2}} \rightarrow\mathbb{R}^{N_{y}}$, where $\mathbf{w}_1$, $\mathbf{w}_2$ are the vectors of learnable parameters. The neural network predicts the output $\mathbf{y}=\mathbf{F}_2(\mathbf{w}_2,\mathbf{F}_1(\mathbf{w}_1,\mathbf{x}_1))$ from the input $\mathbf{x}_1$. We aim to minimize a scalar loss function $L(\mathbf{y}): \mathbb{R}^{N_{y}} \rightarrow\mathbb{R}$, which could for example be a discrepancy measure to some training data or to a prescribed set point curve in MPC. To calculate a corresponding gradient descent update step we need the gradients of the loss with respect to the vectors of trainable parameters. Application of the chain rule yields,
\begin{align}
	\overbrace{\frac{\partial L}{\partial \mathbf{w}_1}}^{[N_{w1}\times 1]} &=&
	\frac{\partial L}{\partial \mathbf{y}}
	\frac{\partial \mathbf{y}}{\partial \mathbf{x}_2}
	\frac{\partial \mathbf{x}_2}{\partial \mathbf{w}_1}&=&
	\overbrace{\boldsymbol{\mathcal{J}}_{x_2,w_1}^\intercal}^{[N_{w1}\times N_{x2}]} 
	\overbrace{\boldsymbol{\mathcal{J}}_{y,x_2}^\intercal}^{[N_{x2}\times N_y]}
	\overbrace{\frac{\partial L}{\partial \mathbf{y}}}^{[N_{y}\times 1]},\\
	\overbrace{\frac{\partial L}{\partial \mathbf{w}_2}}^{[N_{w2}\times 1]} &=&
	\frac{\partial L}{\partial \mathbf{y}}
	\frac{\partial \mathbf{y}}{\partial \mathbf{w}_2}&=&
	\overbrace{\boldsymbol{\mathcal{J}}_{y,w_2}^\intercal}^{[N_{w2}\times N_{y}]} 
	\overbrace{\frac{\partial L}{\partial \mathbf{y}}}^{[N_{y}\times 1]},
\end{align}
where $\boldsymbol{\mathcal{J}}_{\mathbf{a},\mathbf{b}}$ are Jacobian matrices of $\mathbf{a}$ with respect to $\mathbf{b}$, i.e. $(\boldsymbol{\mathcal{J}}_{\mathbf{a},\mathbf{b}})_{i,j} = \frac{\partial \mathbf{a}_i}{\partial \mathbf{b}_j}$. Based on the loss gradients and a learning rate, $\eta$, an update step in the trainable parameters is calculated,
\begin{align}
	\Delta \mathbf{w}_1 = -\eta \frac{\partial L}{\partial \mathbf{w}_1}, \label{equ:GD}\\
	\Delta \mathbf{w}_2 = -\eta \frac{\partial L}{\partial \mathbf{w}_2}.\label{equ:GD2}
\end{align}
After applying the updates, another forward and backward pass is calculated to determine the next update. This is repeated until a sufficiently optimal solution is found. While gradient descent is used here as an example, of course any other more advanced method from RProp \citep{Riedmiller1992RPROPA,Igel2000ImprovingTR} to AdamW \citep{Loshchilov2017} can be employed.\par 

While the backpropagation method is typically selected because of its scalability to large problems, another important aspect is its decentral nature. Every layer and the loss can be regarded as local entities in Figure \ref{fig:backpropagation}. Backpropagation can be implemented solely by communicating loss gradients between these entities, and the content only refers to state variables directly known to the layer as input or output. When receiving a loss gradient each layer uses its Jacobians to translate the incoming gradient to a gradient with respect to its parameter vector, $\frac{\partial L}{\partial \mathbf{w}_i}$, and a gradient with respect to its input data. The gradient with respect to its parameter vector is used to update the latter. The gradient with respect to the input data is communicated to the layer the input was received from. Thus none of the layer's Jacobians, which describe its behaviors and dependencies, are communicated outside. The decentral nature, hence, also adds a certain degree of information privacy, as no model or dependency needs to be explicitly communicated. Nonetheless, a global cost function is optimized by these decentral and possibly unsynchronized actions. \par

While Equation \eqref{equ:GD} is applied as stochastic gradient descent also to non-convex problems in deep-learning by using noisy gradient (subgradient) estimates from subsets of the training data \citep{bottou-bousquet-2008}, we limit this work to convex problems. Whether the noise present in the gradients is suitable, is to be discussed in future research.
\subsection{Differentiable Data Fusion}
To implement the quasi neural network idea for a process chain, and conduct backpropagation, the forward vector field functions of each process/node need to be defined. In a purely digital feed-forward process chain either the processes themselves are differentiable or differentiable models thereof can be derived, that represent these functions. Input data is always available and precise. In a physical process chain where material streams are processed this is not the case. The input and output material streams need to be digitized by noisy sensors. Often industrial sensors cannot capture all properties important to the process, thus requiring additional information from neighboring process nodes. Sensor data may have a too limited sampling rate, and just like communication with neighboring nodes might fail. The chosen method thus needs to be robust to missing information and dynamically able to integrate all information sources.\par 
The problem can be described as a state estimation, where the sought state vectors comprise the necessary input and output material properties, machine parameters, and internal process state (wear, fill level etc.). 
To estimate these local state vectors, the processes which only capture part of their state, can communicate their knowledge to each other, and apply data fusion methods to reach a consensus on exchanged material properties and a lower uncertainty on their own state. These decentral data fusion methods are applied to sensor network applications in environmental monitoring \citep{Jamil2015}, object tracking \citep{Lin2006}, as well as industrial monitoring \citep{Salahshoor2008}. For dynamical systems Kalman filter like predictive models are included to infer time series of states \citep{Salahshoor2008}. The fusion methods typically rely on statistical formulations, be it in the form of Maximum likelihood estimation \citep{Zhou1997}, Bayesian inference \citep{Taylor2019}, or Kalman filtering \citep{Benaskeur}. All result in similar algorithms that can typically be transformed to one another. Hence, without limiting generality we rely on the Bayesian formulation. To keep the inference problems numerically tractable in the later iterative optimization setting, we use multivariate normal (MVN) probability models for all state vectors and observations. Thus, the data fusion result for the state vectors at each node is quantified by the MAP estimate (mean vector) and the posterior covariance matrix.\par 
The MAP estimates represent the physical state of the process chain virtually. The material properties transported by the physical material flow and the causal relationships are represented by the information exchange between nodes and process/prediction models in the inference respectively. Hence, the local data fusion is the sought forward vector field function in the view of a quasi neural network. The inputs to the function are the information pieces fused: process model, prediction model, prediction from the previous time step, parameters, sensor data, communication from other nodes. The output is the MAP state estimate, which is communicated on to other nodes and can be used in the loss function to be optimized. As explained in the previous section, we need the Jacobians of each node/layer to compute a parameter update, i.e. the data fusion process needs to be differentiable.\par 
In the following section we will first introduce the statistical data fusion problem for uncorrelated data, and then introduce common methods to treat unknown correlations. Afterwards, we derive Jacobians for the data fusion problem achieving the differentiable data fusion necessary for gradient descent optimization.

\subsubsection{Statistical Data Fusion}

\begin{figure}[h!]
	\centering
	\includegraphics{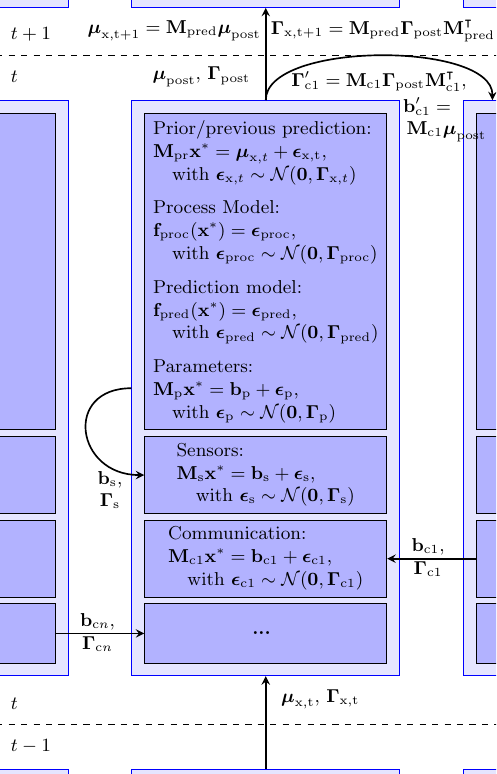}
	\caption{Bayesian inference, for distributed data fusion and prediction.}
	\label{fig:Bayesian_prediction}
\end{figure}

Following ideas of generalized Kalman filtering and MPC, the next time step's state vector is predicted from the current, thus, sequentially predicting a chain of state vectors. Unlike Kalman filtering the prediction is not performed as a separate step but integrated into the actual data fusion in order to regard the time step as a monolithic node in backpropagation. Hence, the data fusion is conducted on a concatenated vector 
\begin{equation}
	\mathbf{x}^* = \begin{bmatrix}
		\mathbf{x}_{t+1}\\
		\mathbf{x}_{t}
	\end{bmatrix},
\end{equation}
where $\mathbf{x}_{t}\in \mathbb{R}^{n_x}$ is the state vector of the current time step, and  $\mathbf{x}_{t+1}\in \mathbb{R}^{n_x}$ refers to the state vector of the next time step with a time step size of $\Delta_t$. Both state vectors share the same structure with ingoing and outgoing material properties, process parameters, and internal process state descriptions. \par 
In Bayesian inference the knowledge about a vector $\mathbf{x}^*$ is encoded in its probability density function (PDF) $p(\mathbf{x}^*)$. 
The data fusion problem for one time step can then be described by Bayes' equation,
\begin{equation}
	p(\mathbf{x}^*|\mathbf{b}) = \frac{p(\mathbf{b}|\mathbf{x}^*)p(\mathbf{x}^*)}{p(\mathbf{b})},
\end{equation}
where $p(\mathbf{b}|\mathbf{x}^*)$ is the likelihood given available sensor data, communications, process and prediction models. $p(\mathbf{x}^*)$ is the prior knowledge available from the previous time step. The evidence $p(\mathbf{b})$ is effectively a scaling constant, and $p(\mathbf{x}^*|\mathbf{b})$ is the posterior distribution describing the total knowledge after data fusion.\par 
From the previous time steps prediction, or an initial estimate, we can define a prior $p(\mathbf{x})$. As previously mentionned we assume MVN PDFs, fully qualified by mean vector and covariance matrix. Our prior information from the previous time step's prediction only concerns the current state, $\mathbf{x}_t$, making the prior degenerate, as information on the prediction $\mathbf{x}_{t+1}$ is missing. However, in classic Bayesian inference a measure proportional to the PDF suffices as long as the likelihood supplies the missing information,
\begin{align}
	p(\mathbf{x})\propto& \exp\left[-\frac{1}{2} (\mathbf{M}_\mathrm{pr}\mathbf{x}^*-\boldsymbol\mu_{\mathrm{x},t})^\intercal\boldsymbol\Gamma_{\mathrm{x},t}^{-1}(\mathbf{M}_\mathrm{pr}\mathbf{x}^*-\boldsymbol\mu_{\mathrm{x},t}) \right],\nonumber \\
	&\textnormal{with } \mathbf{M}_\mathrm{pr}=\left[\mathbf{0}_{n_x\times n_x},\mathbf{I}_{n_x\times n_x}\right].
\end{align}
Here $\boldsymbol\mu_{\mathrm{x},t}$ and $\boldsymbol\Gamma_{\mathrm{x},t}$ are the predicted mean vector and covariance matrix for the current state, $\mathbf{0}_{n_x\times n_x}$ is an $n_x\times n_x$ matrix of zeros and $\mathbf{I}_{n_x\times n_x}$ is an $n_x\times n_x$ identity matrix. The observation matrix $\mathbf{M}_\mathrm{pr}$ projects the concatenated state vector onto the current state vector. The prior can equivalently be expressed by the stochastic equation system
\begin{equation}
	\mathbf{M}_\mathrm{pr}\mathbf{x}^*=\boldsymbol\mu_{\mathrm{x},t} + \boldsymbol\epsilon_\mathrm{pr} \textnormal{, with } \boldsymbol\epsilon_\mathrm{pr}\sim \mathcal{N}(\mathbf{0},\boldsymbol\Gamma_{\mathrm{x},t}),
\end{equation}
where the random vector $\boldsymbol\epsilon_\mathrm{pr}$ describes the uncertainty in the prior, and follows an MVN distribution, $\mathcal{N}(\mathbf{0},\boldsymbol\Gamma_{\mathrm{x},t})$, with mean vector $\mathbf{0}$ and covariance matrix $\boldsymbol\Gamma_{\mathrm{x},t}$.\par 
The likelihood comprises different information sources which, if statistically independent, can be summarized,
\begin{align}
	p(\mathbf{b}|\mathbf{x}^*) =& p(f_\mathrm{proc},f_\mathrm{pred},b_1,b_2,\dots|\mathbf{x}^*) =\\ =&p(f_\mathrm{proc}|\mathbf{x}^*)p(f_\mathrm{pred}|\mathbf{x}^*)p(b_1|\mathbf{x}^*)p(b_2|\mathbf{x}^*)\dots,\nonumber
\end{align}
with the process model, $f_\mathrm{proc}$, prediction model, $f_\mathrm{pred}$, parameter settings, $b_\mathrm{p}$, sensor readings, $b_\mathrm{p}$, and information communicated from other nodes, $b_{\mathrm{c}i}$. All information components can be described as stochastic equation systems as well. The process model is meant to describe constraints imposed on the state vector of a single time step if available, like for example mass conservation. The generally nonlinear relations are described as
\begin{equation}
	\mathbf{f}_\mathrm{proc}(\mathbf{x}^*)=\boldsymbol\epsilon_\mathrm{proc}\textnormal{, with } \boldsymbol\epsilon_\mathrm{proc}\sim \mathcal{N}(\mathbf{0},\boldsymbol\Gamma_{\mathrm{proc}}),
	\label{equ:impl_proc_model}
\end{equation}
where $\mathbf{f}_\mathrm{proc}(\mathbf{x}^*)\in \mathbb{R}^{n_\mathrm{proc}}$ is the implicit description of said constraints, and $\boldsymbol\epsilon_\mathrm{proc}$ describes the constraints' or model uncertainty. Analogously the prediction model $\mathbf{f}_\mathrm{pred}(\mathbf{x}^*)\in \mathbb{R}^{n_\mathrm{pred}}$ together with the prediction model uncertainty $\boldsymbol\epsilon_\mathrm{pred}$ implicitly describes the relation between the current state and the predicted state of the next time step. In Kalman filtering typically the prediction is expressed explicitly via $\mathbf{x}_{t+1}= \mathbf{f}_\mathrm{pred}^\prime(\mathbf{x}_{t})$. This can be translated to the implicit form, analogous to \eqref{equ:impl_proc_model}, via $\mathbf{f}_\mathrm{pred}(\mathbf{x}^*) = \mathbf{f}_\mathrm{pred}^\prime(\mathbf{x}_{t})-\mathbf{x}_{t+1}$.  The models $\mathbf{f}_\mathrm{pred}$ and $\mathbf{f}_\mathrm{proc}$ may be any kind of white-, grey- or black-box model. In particular, they can be neural networks that will be continuously fine-tuned by the ANT. \par 
The vectors $\mathbf{b}_1$, $\mathbf{b}_2$, ... are (partial) observations of the current state vector arising from sensor readings, parameter settings, and communications with other nodes. Their relation to the concatenated state is described by an observations matrix $\mathbf{M}_{\mathrm{b}i}$ in the respective stochastic measurement model, similar to the formulation of the prior,
\begin{equation}
	\mathbf{M}_{\mathrm{b}i}\mathbf{x}^* = \mathbf{b}_i +\boldsymbol\epsilon_\mathrm{\mathrm{b}i}\textnormal{, with } \boldsymbol\epsilon_\mathrm{\mathrm{b}i}\sim \mathcal{N}(\mathbf{0},\boldsymbol\Gamma_{\mathrm{\mathrm{b}i}}),
\end{equation}
Note that for statistical inference the causal classification into cause or effect of setting a parameter or sensing a state is irrelevant, and both are regarded as information contributions to the inference. All information contributions to the data fusion are summarized as their respective stochastic equation systems in Figure \ref{fig:Bayesian_prediction}.\par
The result of the data fusion is encoded in the posterior distribution, which is also an MVN distribution if all priors and likelihoods are MVN distributed and only linear models are used. Here process and prediction models are generally nonlinear, making the assumption of an MVN posterior a necessary approximation in order to achieve compatibility with data fusions in future time steps. Hence, solving the MVN data fusion problem amounts to determining the posterior mean vector, equivalent to the MAP estimate, and the approximate posterior covariance matrix. As well known from literature, the MAP estimate emerging from all MVN data sources can be described by a least squares problem,
\begin{align}
	\boldsymbol\mu_\mathrm{post}=\mathbf{x}_\mathrm{MAP}^*= \qquad\qquad\qquad\qquad\qquad\qquad\nonumber\\
	=\arg\min_{\mathbf{x}^*} \left\lVert\left[
	\begin{matrix}
		\mathbf{L}_\mathrm{pr} \mathbf{M}_\mathrm{pr}\mathbf{x}^*\\
		\mathbf{L}_\mathrm{proc}\mathbf{f}_\mathrm{proc}(\mathbf{x}^*)\\
		\mathbf{L}_\mathrm{pred}\mathbf{f}_\mathrm{pred}(\mathbf{x}^*)\\
		\mathbf{L}_\mathrm{p} \mathbf{M}_\mathrm{p}\mathbf{x}^*\\
		\mathbf{L}_\mathrm{s} \mathbf{M}_\mathrm{s}\mathbf{x}^*\\
		\mathbf{L}_{\mathrm{c}1} \mathbf{M}_{\mathrm{c}1}\mathbf{x}^*\\
		\dots
	\end{matrix}\right]-\left[
	\begin{matrix}
		\mathbf{L}_\mathrm{pr}\boldsymbol\mu_{\mathrm{x},t}\\
		\mathbf{L}_\mathrm{proc}\mathbf{0} \\
		\mathbf{L}_\mathrm{pred}\mathbf{0} \\
		\mathbf{L}_\mathrm{p}\mathbf{b}_\mathrm{p}\\
		\mathbf{L}_\mathrm{s}\mathbf{b}_\mathrm{s}\\
		\mathbf{L}_{\mathrm{c}1}\mathbf{b}_{\mathrm{c}1}\\
		\dots
	\end{matrix}\right]\right\rVert_2^2,\label{equ:MAP_nonlin}
\end{align}
where $\mathbf{L}_\square$ are the whitening operators with respect to the corresponding covariances $\boldsymbol\Gamma_\square=(\mathbf{L}_\square^\intercal\mathbf{L}_\square)^{-1}$. To estimate the posterior covariance the generally nonlinear process and prediction models are linearized at $\mathbf{x}_\mathrm{MAP}$ using their Jacobians $\mathcal{J}_{f\mathrm{proc},\mathbf{x}^*}$, and $\mathcal{J}_{f\mathrm{pred},\mathbf{x}^*}$, yielding \citep{Jackson1979,Tarantola1982,Kaipio2004},
\begin{align}
	\boldsymbol\Gamma_\mathrm{post} =& (\mathbf{M}_\mathrm{pr}^\intercal\Gamma_{\mathrm{x},t}^{-1}\mathbf{M}_\mathrm{pr} \label{equ:post_cov}\\
	&+\mathcal{J}_{f\mathrm{proc},\mathbf{x}^*}^{\intercal} \boldsymbol\Gamma_{\mathrm{proc}}^{-1} \mathcal{J}_{f\mathrm{proc},\mathbf{x}^*}\nonumber\\
	&+\mathcal{J}_{f\mathrm{pred},\mathbf{x}^*}^{\intercal} \boldsymbol\Gamma_{\mathrm{pred}}^{-1} \mathcal{J}_{f\mathrm{pred},\mathbf{x}^*}\nonumber\\
	& + \mathbf{M}_\mathrm{\mathrm{p}}^\intercal\Gamma_\mathrm{p}^{-1}\mathbf{M}_\mathrm{p}\nonumber\\
	& + \mathbf{M}_\mathrm{\mathrm{s}}^\intercal\Gamma_\mathrm{s}^{-1}\mathbf{M}_\mathrm{s}+\dots)^{-1},\nonumber
\end{align}
where each of the summands describes the information matrix contributed by the respective information source.
The information on the predicted state for the next time step can be cut from $\boldsymbol\mu_\mathrm{post}$ and $\boldsymbol\Gamma_\mathrm{post}$,
\begin{align}
	\boldsymbol\mu_{\mathrm{x},t+1}=& \mathbf{M}_\mathrm{pred}\boldsymbol\mu_\mathrm{post},
	\\
	\boldsymbol\Gamma_{\mathrm{x},t+1}=& \mathbf{M}_\mathrm{pred}\boldsymbol\Gamma_\mathrm{post}\mathbf{M}_\mathrm{pred}^\intercal,\\
	\textnormal{with } \mathbf{M}_\mathrm{pred}=&\left[\mathbf{I}_{n_x\times n_x},\mathbf{0}_{n_x\times n_x}\right],
\end{align}
and is used in the next time steps inference as prior knowledge. Similarly, the augmented knowledge state is communicated back to neighboring nodes, like for example,
\begin{align}
	\boldsymbol\mu_\mathrm{c1}^\prime=& \mathbf{M}_\mathrm{c1}\boldsymbol\mu_\mathrm{post},
	\\
	\boldsymbol\Gamma_\mathrm{c1}=& \mathbf{M}_\mathrm{c1}\boldsymbol\Gamma_\mathrm{post}\mathbf{M}_\mathrm{c1}^\intercal.
\end{align}
These communications are illustrated in Figure \ref{fig:Bayesian_prediction}. A detailed discussion of the communication algorithm is presented at the end of section \ref{sec:methods}.

\subsubsection{Covariance Intersection}
The communication between nodes and the ensuing fusion into new predictions necessarily introduces correlations between the state estimates of nodes throughout the process network. All following communication between nodes, hence, has to be assumed to contain correlated information \citep{MartinLiggins2017}. These correlations need to be regarded in each local inference in order to provide trustworthy results. However,  tracking the exact values of these correlations through the network is practically infeasible. Instead, methods are available from literature to treat inference problems in which the information sources are correlated to an unknown extent. For MVN estimates an often used method is Covariance Intersection (CI) \citep{Julier1997,Hurley,MartinLiggins2017}, which derives a rule for guaranteed conservative estimates of posterior covariance and mean. Based on CI other intersection based methods have been proposed like Ellipsoidal Intersection \citep{Sijs2012} and Inverse Covariance Intersection (ICI) \citep{Noack2017} which promise tighter bounded posterior covariances, but do not guarantee conservative results for all cases \citep{Noack2017a}. Accordingly for its simple implementation and conservatism we choose CI in this work. The following paragraphs explain CI and adapts it to our problem.\par 
Given two state estimates $(\boldsymbol\mu_1,\boldsymbol\Gamma_1)$, $(\boldsymbol\mu_2,\boldsymbol\Gamma_2)$ with unknown cross-covariance $\boldsymbol\Gamma_{1,2}$, a conservative new state estimate is to be calculated. Independent of the unknown cross-covariance a consistent upper bound for the posterior covariance can be constructed from a convex combination of inverse covariances \citep{Julier1997},
\begin{align}
	\boldsymbol\mu_\mathrm{post}=& \boldsymbol\Gamma_\mathrm{post}(w\boldsymbol\Gamma_1^{-1}\boldsymbol\mu_1+(1-w)\boldsymbol\Gamma_2^{-1}\boldsymbol\mu_2)\\
	\boldsymbol\Gamma_\mathrm{post}=& (w\boldsymbol\Gamma_1^{-1}+(1-w)\boldsymbol\Gamma_2^{-1})^{-1}
\end{align}
with the weight $0\leq w\leq 1$. For all valid $w$ the posterior estimate is consistent, but for non-divergent estimates the weight needs to be chosen according to an optimality criterion, like the minimum determinant of $\boldsymbol\Gamma_\mathrm{post}$ \citep{Julier1997} as used in this work. The resulting upper bound is used as the fused estimate in CI, and is proven to be tight \citep{Reinhardt2015}. The CI fusion rule can be consistently extended to fusing $n$ estimates \citep{Niehsen}, thereby losing its tightness \citep{Ajgl2018},
\begin{align}
	\boldsymbol\mu_\mathrm{post}=&  \boldsymbol\Gamma_\mathrm{post}\left(\sum_i^n w_i\boldsymbol\Gamma_i^{-1}\boldsymbol\mu_i\right),\label{equ:CI}\\
	\boldsymbol\Gamma_\mathrm{post}=& \left(\sum_i^n w_i\boldsymbol\Gamma_i^{-1}\right)^{-1},\\
	\textnormal{with } &  w_i \geq 0 \\
	\textnormal{and }& \sum_i w_i =1.
\end{align}
Put differently, CI weights the information matrices in equation \eqref{equ:post_cov} by the corresponding $w_i$, respectively weights the whitening matrices in equation \eqref{equ:MAP_nonlin} by the corresponding $\sqrt{w_i}$.\par 
The information packages grouped in Figure \ref{fig:Bayesian_prediction} are defined according to their source and meaning, not according to their correlation structure, meaning the general weighting described in Equation \eqref{equ:CI} is not appropriate. For example the uncertainty in prior, process model, internal sensor readings and parameter settings is uncorrelated, hence, they can be treated with a single weight factor $w_\mathrm{local}$. The communicated information from other nodes on the other hand must always be assumed to be correlated with the local information and any other communication because of possible previous communication in the current or any previous time step. The prediction model on the other hand can be assumed absolutely uncorrelated with other information packages as it only imposes restrictions on the prediction and no backward information flow from future time steps is regarded here. It could thus be fused with the other information packages in a separate inference without CI. The resulting CI weighting association can be summarized as in Table \ref{tab:CI_weights} with the following conditions on the non-negative weights,
\begin{align}
	w_\mathrm{local} + \sum_i w_{\mathrm{c}i} =& 1,\\
	w_\mathrm{pred} =&1.
\end{align}
\begin{table}
	\caption{CI weight distribution.}
	\begin{tabular}{r|c|c|c|c|c|c|c}
		\textbf{Cov.} & $\boldsymbol\Gamma_{\mathrm{x},t}$ & $\boldsymbol\Gamma_{\mathrm{proc}}$ & $\boldsymbol\Gamma_{\mathrm{s}}$ & $\boldsymbol\Gamma_{\mathrm{p}}$& $\boldsymbol\Gamma_{\mathrm{pred}}$ & $\boldsymbol\Gamma_{\mathrm{c1}}$ & $\dots$\\
		\hline
		\hline
		\textbf{Weight} &\multicolumn{4}{|c|}{$w_\mathrm{local}$} & $w_\mathrm{pred}$ & $w_\mathrm{c1}$ & $\dots$  
	\end{tabular}
	\label{tab:CI_weights}
\end{table}
The weights are chosen to minimize the determinant of the posterior covariance in equation \eqref{equ:post_cov}. As the Jacobians of the generally nonlinear models $\mathcal{J}_{f\mathrm{proc},\mathbf{x}^*}$ and $\mathcal{J}_{f\mathrm{pred},\mathbf{x}^*}$, are dependent on the fusion result $\mathbf{x}_\mathrm{MAP}^*$, their working point is approximated by first performing a naive fusion with all CI weights set to one. Afterwards the data fusion is conducted with the final CI weights, and the MAP state estimate represents the current state of knowledge. In order to optimize this state in favor of a quality criterion the relation between input information, $\boldsymbol\mu_{\mathrm{x},t}$, $\mathbf{b}$, and the output, $\mathbf{x}_\mathrm{MAP}^*$, needs to be known, which is discussed in the next section.

\subsubsection{MAP derivatives and their use in backpropagation}
Equation \eqref{equ:MAP_nonlin} (including CI weighting) implicitly assigns a value, $\mathbf{x}_\mathrm{MAP}^*$, to every set of input data, models, $\boldsymbol\mu_{\mathrm{x},t}$, $\mathbf{b}$, and can be shortened to,
\begin{align}
	\mathbf{x}^*_\mathrm{MAP} =& \arg\min\limits_{\mathbf{x}^*} \left\lVert  \mathbf{L}\mathbf{F}(\mathbf{x}^*) - \mathbf{L}\boldsymbol\beta \right\rVert_2^2 \nonumber\\
	=& \arg\min\limits_{\mathbf{x}^*} \left\lVert \mathbf{f}(\mathbf{x}^*,\boldsymbol\beta) \right\rVert_2^2\nonumber\\
	=& \arg\min\limits_{\mathbf{x}^*}  (\mathbf{f}(\mathbf{x}^*,\boldsymbol\beta))^T \mathbf{f}(\mathbf{x}^*,\boldsymbol\beta) \nonumber\\
	=& \arg\min\limits_{\mathbf{x}^*} \chi^2(\mathbf{x}^*,\boldsymbol\beta), 
	\label{equ:minim_simplified}
\end{align}
where $\mathbf{L}$ is the block-diagonal whitening matrix comprised of all other whitening matrices, $\mathbf{F}(\mathbf{x}^*)$ is the concatenation of the process model, the prediction model, and the linear observation models of prior, sensors, parameters, and communications, $\boldsymbol\beta$ is the concatenated vector of all inhomogeneities/mean values (right side of the equation systems in Figure \ref{fig:Bayesian_prediction}: measurement values, communicated mean values, prediction from previous state, zero vectors for the models). The resulting vector difference is expressed as vector-valued function $\mathbf{f}(\mathbf{x}^*,\boldsymbol\beta)$, and the sum of squares of this vector function is $\chi^2(\mathbf{x}^*,\boldsymbol\beta)$.\par 
When viewing just the vicinity of a minimum $\mathbf{x}^*_\mathrm{MAP}$, the minimization problem in equation \eqref{equ:minim_simplified} can be reduced to the equation system describing the property of an extremum, that is the derivatives of $\chi^2$ with respect to minimization parameters need to be zero,
\begin{equation}
	\label{equ:extremum_eqns}
	\frac{\partial\chi^2}{\partial \mathbf{x}^*}(\mathbf{x}^*,\boldsymbol\beta) =\mathbf{k}(\mathbf{x}^*,\boldsymbol\beta)= \mathbf{0}.
\end{equation}

The equation system \eqref{equ:extremum_eqns} essentially assigns a vector $\mathbf{x}_\mathrm{MAP}^*$ for each inhomogeneity vector $\boldsymbol\beta$. Put differently, this nonlinear equation system is the implicit description of a function, $\mathbf{x}_\mathrm{MAP}^*=g(\boldsymbol\beta)$, in the vicinity of the actual $\mathbf{x}_\mathrm{MAP}^*$. The sought derivatives of the state vector are thus derivatives of the implicit function, $g(\boldsymbol\beta)$, summarized in the Jacobian, $\mathcal{J}_{g, \boldsymbol\beta}$. 
Assuming $\mathbf{k}$ is continuously differentiable and $\mathcal{J}_\mathbf{k,x}$ is invertible, the implicit function theorem allows to calculate this Jacobian. Neglecting second order terms in $\mathbf{f}$ yields, 
\begin{align}
	\mathcal{J}_{\mathbf{g},\boldsymbol\beta} \approx& -\left[ \left(\frac{\partial\mathbf{f}}{\partial \mathbf{x}^*}\right)^T\frac{\partial\mathbf{f}}{\partial \mathbf{x}^*} \right]^{-1} \left(\frac{\partial\mathbf{f}}{\partial \mathbf{x}^*}\right)^T\frac{\partial\mathbf{f}}{\partial \boldsymbol\beta}=\nonumber\\
	=& \left[ \left(\mathbf{L}\mathcal{J}_{\mathbf{F},\mathbf{x}^*}\right)^T\mathbf{L}\mathcal{J}_{\mathbf{F},\mathbf{x}^*} \right]^{-1} \left(\mathbf{L}\mathcal{J}_{\mathbf{F},\mathbf{x}^*}\right)^T\mathbf{L}
\end{align}
For a detailed derivation see Appendix \ref{app:implderiv}. This Jacobian represents the derivatives of the device state with respect to $\boldsymbol\beta$ which contains optimizable parameters, information from neighboring nodes, the prior mean from the previous prediction and model inhomogeneities. The Jacobian can therefore be cut into different chunks of rows, each referring to a different information source: $\mathcal{J}_{\mathbf{g},\mu_{\mathrm{x},t}}$, $\mathcal{J}_{\mathbf{g},\mu_{\mathrm{proc}}}$, $\mathcal{J}_{\mathbf{g},\mu_{\mathrm{pred}}}$, $\mathcal{J}_{\mathbf{g},\mathbf{b}_{\mathrm{p}}}$, $\mathcal{J}_{\mathbf{g},\mathbf{b}_{\mathrm{s}}}$, $\mathcal{J}_{\mathbf{g},\mathbf{b}_{\mathrm{c}1}}$. Similar to the neural network in Figure \ref{fig:backpropagation} these Jacobian components are used to backpropagate to other neighbors, via  $\mathcal{J}_{\mathbf{g},\mathbf{b}_{\mathrm{c}1}}$, to previous time steps,  via  $\mathcal{J}_{\mathbf{g},\mu_{\mathrm{pred}}}$, or to the parameters to be optimized, via $\mathcal{J}_{\mathbf{g},\mathbf{b}_{\mathrm{p}}}$. Different from the simple feed forward network in Figure \ref{fig:backpropagation} the ANT might feature a bi-directional or even circular communication graph for information exchange as well as backpropagation. A parameter therefore might receive multiple gradients that need to be summed before calculating an update step according to equation \eqref{equ:GD}. \par 
In Figure \ref{fig:backpropagation} always the full output of a layer - here, the state estimate - is communicated to the next layer - here, the neighbor node or next time step. In the ANT, however, only the relevant information is communicated by projecting the state estimate through the observation matrices, $\mathbf{M}_\mathrm{pred}$, and $\mathbf{M}_{\mathrm{c}i}$. Hence, if a loss gradient is received from a future time step it has the same dimensions as $\boldsymbol\mu_{\mathrm{x},t+1}$, and is translated to reference the full MAP vector, $\mathbf{x}^*_\mathrm{MAP}=\boldsymbol{\mu}_\mathrm{post}$, 
\begin{equation}
	\frac{\partial L}{\partial  \mathbf{x}^*_\mathrm{MAP}} = \mathbf{M}_\mathrm{pred}^\intercal \frac{\partial L}{\partial \boldsymbol\mu_{\mathrm{x},t+1}}.
\end{equation}
Analogously for a gradient received form a neighbor, the gradient is translated,
\begin{equation}
	\frac{\partial L}{\partial  \mathbf{x}^*_\mathrm{MAP}} = \mathbf{M}_{\mathrm{c}i}^\intercal \frac{\partial L}{\partial \mathbf{b}_{\mathrm{c}i}}.
\end{equation}
The loss gradient with respect to the MAP is then translated to loss gradients with respect to the input quantities of the data fusion, that are,
\begin{itemize}
	\item $\frac{\partial L}{\partial  \mathbf{x}^*_{t}}   =  \mathcal{J}_{\mathbf{g},\boldsymbol{\mu}_{x,t}}^\intercal  \frac{\partial L}{\partial  \mathbf{x}_{\mathrm{MAP}}}$, which is backpropagated to the previous time step,
	\item $\frac{\partial L}{\partial  \mathbf{b}_{\mathrm{c}i}}   =  \mathcal{J}_{\mathbf{g},\mathbf{b}_{\mathrm{c}i}}^\intercal  \frac{\partial L}{\partial  \mathbf{x}_{\mathrm{MAP}}}$, which is agglomerated in the gradient storage of the respective information set, and backpropagated to the corresponding neighbor,
	\item $\frac{\partial L}{\partial  \mathbf{b}_{\mathrm{b}\mathrm{p}}}   =  \mathcal{J}_{\mathbf{g},{\mathrm{b}\mathrm{p}}}^\intercal  \frac{\partial L}{\partial  \mathbf{x}_{\mathrm{MAP}}}$, which is agglomerated in the gradient storage for the respective parameter and then used to update the parameter analogously to equation \eqref{equ:GD},
	\item $\frac{\partial L}{\partial  \mathbf{b}_{\mathrm{b}\mathrm{s}}}   =  \mathcal{J}_{\mathbf{g},\mathbf{b}_{\mathrm{bs}}}^\intercal  \frac{\partial L}{\partial  \mathbf{x}_{\mathrm{MAP}}}$, which is agglomerated at the sensor for eventual fine tuning of AI-methods in the sensor,
	\item $\frac{\partial L}{\partial  \boldsymbol\mu_{\mathrm{proc}}}   =  \mathcal{J}_{\mathbf{g},\boldsymbol\mu_{\mathrm{proc}}}^\intercal  \frac{\partial L}{\partial  \mathbf{x}_{\mathrm{MAP}}}$, and
	$\frac{\partial L}{\partial  \boldsymbol\mu_{\mathrm{pred}}}   =  \mathcal{J}_{\mathbf{g},\boldsymbol\mu_{\mathrm{pred}}}^\intercal  \frac{\partial L}{\partial  \mathbf{x}_{\mathrm{MAP}}}$, which are agglomerated for eventual fine tuning of AI-based process and prediction models.
\end{itemize}
Thus, loss gradients are only sent to neighbors or time steps from which previously information was received for data fusion. Accordingly, in Figure \ref{fig:gradient_flow} the gradient arrows are always opposed to the information flow. If the information exchange between nodes is two-way, the gradient exchange is also two-way, and both may require a convergence threshold to avoid messaging back and forth indefinitely. This and and the general messaging and processing protocol are described in the next section.
\subsection{ANT algorithm}
\begin{figure}[h!]
	\begin{center}
		\includegraphics[width=8.4cm]{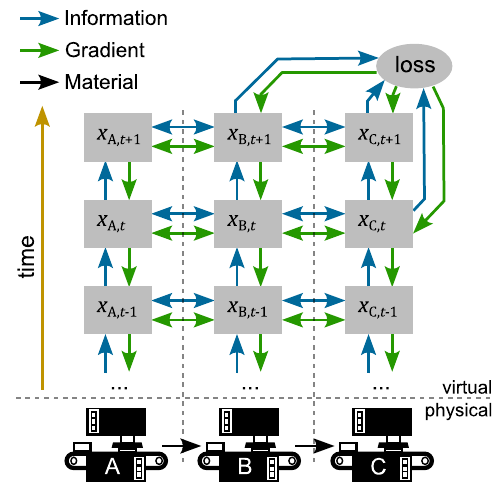}
		\caption{Communication network between time steps and processes, for information exchange and gradient propagation.}
		\label{fig:gradient_flow}                               
	\end{center}                                 
\end{figure}
The two main tasks of the Artificial Neural Twin are decentral inference of the machine states by state information, and, given that state, backpropagating loss gradients in order to optimize process parameters or AI-models. In order to avoid stability issues we chose to discern an information period meant to exchange information and conduct decentral data fusion, and a backpropagation period, during which the state estimates are frozen, in order for backpropagation to happen in a fixed environment. Three message types are exchanged between machines:
\begin{enumerate}
	\item \textbf{information messages} for communicating mean vectors and covariance matrices of state estimates between machines,
	\item \textbf{gradient messages} for backpropagating loss gradient vectors, and
	\item \textbf{control messages} for commands like switching between information and backpropagation period, freezing/unfreezing states, applying an update step or zeroing gradients.
\end{enumerate}
The contents of the messages are described in Table \ref{tab:messages}.\par
\begin{table}
	\caption{Contents of the three types of messages.}
	\label{tab:messages}
	\begin{tabular}{l|ccc}
		\textbf{Type} & \multicolumn{1}{c|}{information} & \multicolumn{1}{c|}{gradient} & control\\
		\hline\hline
		\textbf{Fields} & \multicolumn{3}{c}{sender, recipient, type}\\ \cline{2-4}
		& \multicolumn{2}{c|}{time stamp} & action\\ \cline{2-4}
		& \multicolumn{1}{c|}{mean vector} & \multicolumn{1}{c|}{gradient vect.} & \\ \cline{2-3}
		& \multicolumn{1}{c|}{covariance matr.} & 
	\end{tabular}
\end{table}
The decision on the change between information and backpropagation period is made by a separate node entity, that also takes care of evaluating the MPC loss function and sending the initial loss gradients for backpropagation. It is therefore called loss node and displayed in Figure \ref{fig:gradient_flow}. In order to enable communication  each node must have registered its relevant neighbors by name and corresponding observation matrix, $\mathbf{M}_{\mathrm{c}i}$ which projects the state vector onto an agreed upon format for exchange between the nodes (e.g. only the subsection of the state vector that describes the exchanged material). \par 
The loss node needs to be a registered neighbor of every node, that has relevant information for the loss function. The ANT network is initialized in the information period, meaning all nodes communicate their state estimates and fuse newly received information. Information is only sent if either no previous message has been sent to the neighbor for that time step, or if the Kullback-Leibler (KL) divergence between the already sent information and the new state estimate is larger than the information communication threshold, $\Delta_\mathrm{info}$. Time steps in the history of each node are added if necessary to keep a defined finite horizon lookout into the future. If a time period of $\Delta t_\mathrm{switch}$ has passed and optimization is activated the loss node sends a control message to all nodes, ordering them to zero the gradient storages, freeze the state, and start the backpropagation period. After the control message the loss node evaluates the loss function and sends out the gradient messages to the nodes. This starts a gradient message avalanche as each node backpropagates the received gradients through time and to their neighbors. The number of messages decays when the gradient magnitudes reach the gradient communication threshold, defined relatively by $\Delta_\mathrm{grad}$. The relevant loss gradients with respect to the optimizable parameters are now stored in each nodes gradient storages. A fixed time interval after switching to gradient mode, $\Delta t_\mathrm{switch}$, the loss node sends out another control message for the nodes to apply a gradient descent update step, unfreeze the states (resolve data fusion for all necessary time steps), and communicate information messages again. After another period $\Delta t_\mathrm{switch}$ the next switch to backpropagation happens, thus iteratively finding a consensus and optimizing.\par 
From the view of an ANT node the algorithm is described in a listing in Appendix \ref{app:ant_alg}.

\subsubsection{AI-model fine tuning}
While a detailed discussion or demonstration of AI-model fine tuning through the ANT is beyond the scope of this work, it is nonetheless one of the long term motivations behind the ANT, and, hence, is outlined conceptually in this section.\par 
AI methods are becoming more abundant in industrial applications \citep{schieringmodellbildung,Krauss2018}, be it in process models like in the ANT, or in AI sensory, e.g. deep-learning based preprocessing in camera based sensors. These models initially require a significant investment in training data sets. However, their accuracy in practical applications might decline as the input data distributions drift over long periods of time. A good example can be seen in AI-based sorting methods for used plastic packagings: while initially trained on current data, new packaging types will reach the market over the years, and thus also the sorting plants. The change in distributions would typically require a new investment in a data set, but the ANT promises to be an alternative.\par
Typically despite the best process supervision, quality control through spot checks is necessary. These spot checks can be exploited by the ANT as a continuous source of training stimuli. The spot check results are compared to the predictions of the ANT for the corresponding time step in a regression cost function (e.g. L2-loss). The resulting loss gradients are backpropagated through the ANT network to all AI-models and sensors used as information source in the decentral data fusion. An important difference to the gradient propagation for process optimization is that the MPC loss gradients are only backpropagated to the current time step as it is not reasonable to optimize machine parameters in the past. The regression loss gradients from spots checks on the other hand can only emerge in the current or a past time step. Hence, they backpropagate to all time steps available in past history. After the backpropagation period ends, the gradient storage of each process model and for each sensor measurement contains a corresponding loss gradient, that can be passed on to an external training method specific to the respective AI-model. \par 

Another possibility lies in the use of overdetermined information sources directly available in the ANT network, for example two sensors measureing the same quantities at different locations. A contradiction between observations will lead to an increased $\chi^2(\mathbf{x}^*_\mathrm{MAP},\boldsymbol\beta)$ after data fusion at one node (see Equation \eqref{equ:minim_simplified}). Hence, the training loss function is given by the inference, integrating the retraining concept directly in the general ANT concept.\par

Practicle implementation in both cases requires regarding several issues:
\begin{itemize}
	\item Each single training stimulus is very weak, requiring care when designing the model specific training method.
	\item The process chain is expected to constantly operate close to optimal conditions, thus strongly limiting the distribution of the available data in the stimuli. In order not to overfit the model to this narrow region of working conditions, regularization methods or similar need to be employed (catastrophic forgetting \citep{Parisi2019}).
	\item The AI-model specific external training methods need to keep a history of input data or relevant internal layer Jacobians in order to facilitate further backpropagation into the model.
\end{itemize}
Details of the implementation are subject to future research.

\section{Results}\label{sec:Results}
Plastic recycling is a important topic given the latent concern about the environmental impact of this waste. For this reason, the plastic recycling process was selected, as a test scenario for the capabilities of the ANT. A virtual machine park with three machine types is simulated: a sieving drum, conveyor belts for introducing dead times, and magnetic sorters. Each machine or process step is represented by an ANT node with its accompanying process models and state description. After imposing a loss function dependent on the output material flow, the distributed ANT nodes start optimizing their parameters. This is demonstrated for steady operation as well as a fluctuating material composition.

\subsection{Virtual facility}

Since the idea is to assess the properties of the ANT with respect to ground truth mass flow compositions, which are hard to acquire in real world processes, we implement a section of the sorting process as a virtual facility. We use the game engine environment Unity \citep{Juliani2018}. Our aim is to represent a process chain that if well does not exactly simulate a real process, recreates the problems such a process faces, for instance: recyclable materials falling out of conveyor belts, getting stuck between machine elements, or the inclusion of impurity on the sorting outputs caused by the normal behavior of a machine whose parameters have not been optimized. 

In the mentioned process, we simulated two types of sorting; size and magnetic sorting, separated by material-transporting conveyor belts, for which the time delay is taken into account.

These sorting types therefore require corresponding materials; for which we developed three different sizes of objects (Small, Medium, and Large), split in two ferromagnetic categories, Ferromagnetic (FM) and Non-Ferromagnetic (NFM). The instantiation of recyclable articles into the sorting environment is modeled as a Poisson process.

\begin{figure*}
	\begin{center}
		\includegraphics[scale=0.4]{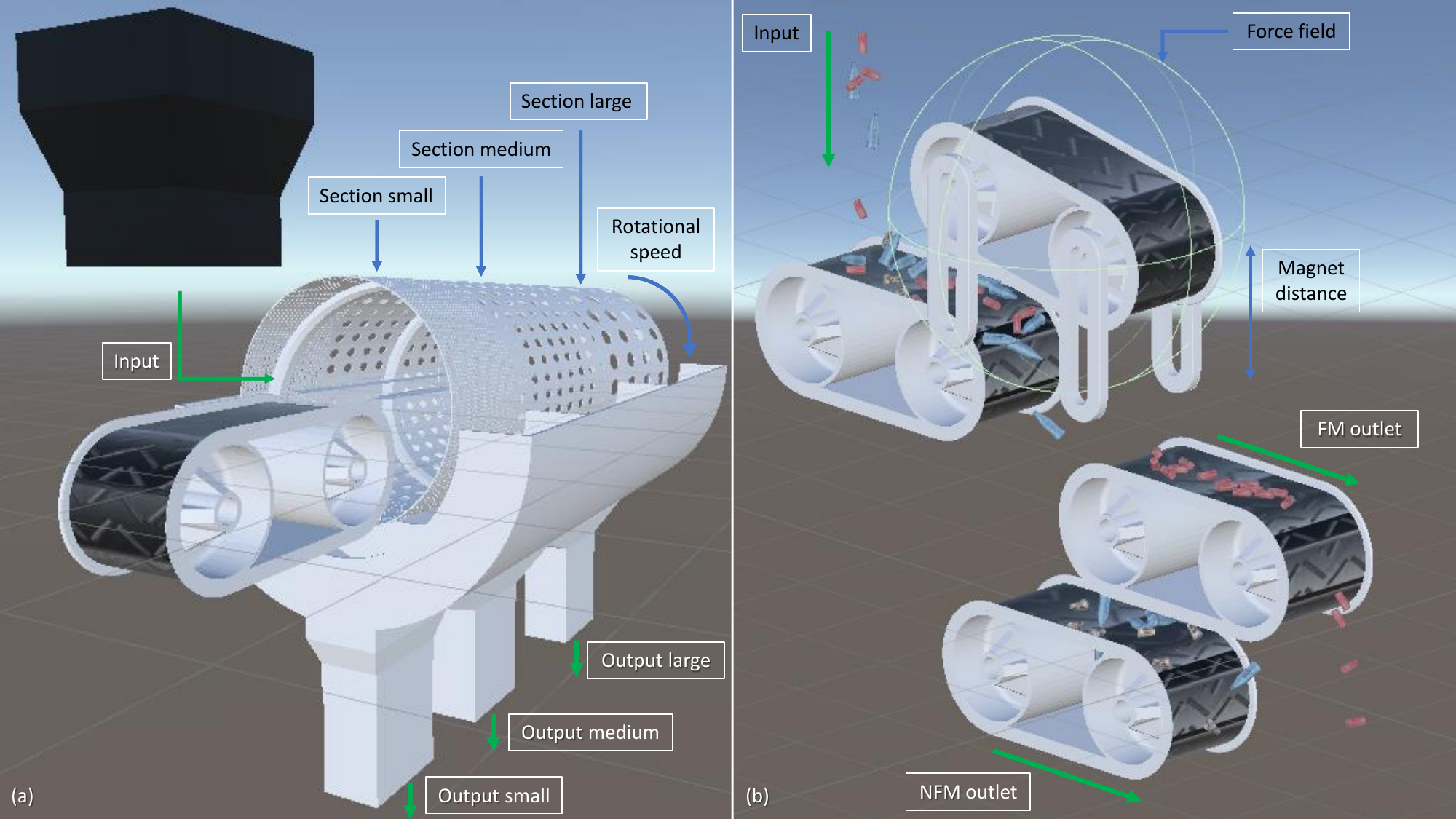}
		\caption{Machine components: Sieving machine (left), showing the rotational speed (optimization parameter), inlet, and outlets for three object sizes. Magnetic sorter (right), showing the force field, magnet distance (optimization parameter), and the outlets for two material types}
		\label{fig:machines_diagram}                             
	\end{center}                               
\end{figure*}

The size sorting is emulated by using a sieving drum which rotates with variable speed (machine optimization parameter). As shown in Figure \ref{fig:machines_diagram}(a), the cylinder walls are divided in three sections, with different hole diameters for each section so that the three object sizes can be sorted.

The magnetic sorting process comprises separating ferromagnetic materials (FM) from the non-ferromagnetic (NFM). In order to do that, we used a force field that is located in the magnet conveyor belt shown in Figure \ref{fig:machines_diagram} on the right. This force field is distance dependent (machine optimization parameter); therefore, the closer the magnet to the FM objects the higher the acceleration acting on them. This acceleration is what separates the FM from the NFM objects and conducts them to their respective machine outlet.

\begin{figure}
	\begin{center}
		\includegraphics[scale=0.38]{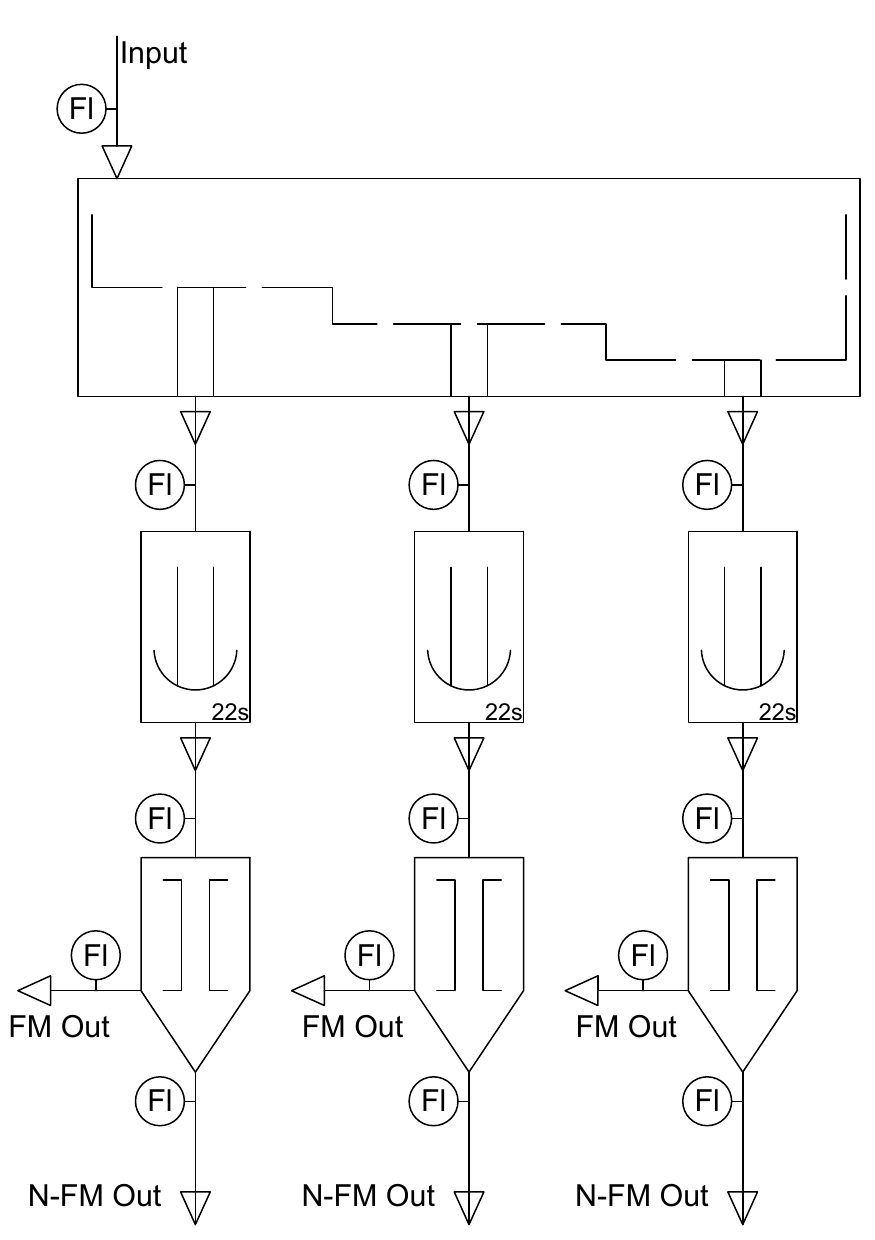}
		\caption{Flow diagram of the process. Input on sieving machine, afterwards three long conveyor belts and at the end three magnetic sorters, with corresponding flow sensors (Fl) in the respective sections of the process.}
		\label{fig:flo_diag}                             
	\end{center}                               
\end{figure}
As figure \ref{fig:flo_diag} shows, the flow process of the virtual facility starts at the top of the drum of the sieving machine, where first the small materials are filtered out, followed by the medium and large on their respective outputs; here, flow sensors measure the material mass flow going though the respective output. Then conveyor belts transport each size fraction to a corresponding magnetic sorter, inducing a dead time of 32~s. The magnetic sorter captures input and output mass flows as well.

A more in detail description of the virtual facility can be found in \ref{Apdx:Virtual_facility}.

\subsubsection{Process models}
Depending on the availability of training data or physical models, the ANT can accommodate black, gray or white box models as long as they are differentiable. These types of models are demonstrated on the sieving machine, the conveyor belt, and the magnetic sorter, which represent individual ANT nodes in this experiment.


In the case of the sieving machine, a gray box model was used. Here the probabilities of an object of a certain size class to exit through each outlet were modeled as linear functions of the drum's rotational speed. On the other hand,  the mixing phenomenon inside the sieving drum as well as the dead time was modeled using a convolution of the past input time series and an impulse response function. The sieving probabilities, mixing and dead times of the model were fitted to data collected in the virtual facility.


As an example of a white box model, we develop a representation of a conveyor belt, where the inputted material is outputted with no variations in mass but with a predetermined dead time of 32~s.

For the magnetic sorting machine a black box model was implemented. Here a feed forward neural network was used, with its characteristic input (Input mass flow, plus magnet distance), hidden, and output layers (true and false positives for FM and NFM outputs respectively). This network was trained on a data set that was generated in the previously mentioned virtual facility, which includes all the features required by the model (Input mass flow, magnet distance, true and false positive FM, true and false positive NFM). To link this material description to the size class description used in the sieving machine and conveyor belt, statistic independence between size and ferromagnetism is postulated (with high model uncertainty).

For more details about the process models, please refer to the Appendix \ref{Apdx:process_models}.

\subsection{ANT-facility data exchange}

The ANT is meant to work side by side the facility as it uses real-time sensor readings in order to infer predictions of future states. 

Inside the virtual facility, article mass counting sensors have been set at the input of the facility and at the output of every represented machine. Figure \ref{fig:flo_diag} depicts these sensors with the "Fl" legend. While fraction specific mass flows are available from the sensors, the ANT nodes are only provided total mass flows. No additional synthetic noise is added to the readings as the approximately 18~\% uncertainty from the Poisson-noise (30~s integration period) is deemed sufficiently large to represent realistic conditions.

The discrete time step size of the ANT nodes is set to $\Delta_t=30$~s. The switching time between information and backpropagation period is set to $\Delta t_\mathrm{switch}=15$~s. On each transition between time steps each node sets the optimized parameter specified by the corresponding time step on the virtual machine.

The finite horizon (time period for inference of new time steps) is set to eight time steps (240~s) on each ANT node. The minimum of information gained to trigger  information exchange between nodes is $\Delta_\mathrm{info}=10^{-6}$~bit, and the gradient magnitude must be at least $\Delta_\mathrm{info}=10^{-6}$ of the last gradient magnitude in order to send gradient messages.

\subsection{Machine performance optimization}

Exemplarily, the facility is optimized for maximum profit as price multiplied with mass flow. The price per kilogram of output material is modeled quadratically dependent on its purity. The purity is defined as the ratio between mass flow correctly sorted in size and ferromagnetic property and the total mass flow of the corresponding output. Applying this scheme to the two outputs of the middle size magnetic sorter yields the loss function,
\begin{align}
	loss =-
	& \left(\frac{\dot{m}_{\mathrm{NFM^{TP}_{M}}}} {\dot{m}_{\mathrm{NFM_{Total}}}}\right)^2 \dot{m}_{\mathrm{NFM^{TP}_{M}}} -\left(\frac{\dot{m}_{\mathrm{FM^{TP}_{M}}}} {\dot{m}_{\mathrm{FM_{Total}}}}\right)^2 \dot{m}_{\mathrm{FM^{TP}_{M}}},\\
	\dot{m}_{\mathrm{NFM_{Total}}} 		    	&=\dot{m}_{\mathrm{NFM_{S}}}+\dot{m}_{\mathrm{NFM_{M}}}+\dot{m}_{\mathrm{NFM_{L}}},\nonumber\\
	\dot{m}_{\mathrm{NFM}_{\textit{p}}}
	&=\dot{m}_{\mathrm{NFM^{TP}_{\textit{p}}}} + \dot{m}_{\mathrm{NFM^{FP}_{\textit{p}}}},\nonumber\\
	\dot{m}_{\mathrm{FM_{Total}}} &=\dot{m}_{\mathrm{FM_{S}}}+\dot{m}_{\mathrm{FM_{M}}}+\dot{m}_{\mathrm{FM_{L}}},\nonumber\\
	\dot{m}_{\mathrm{FM}_{\textit{p}}}
	&=\dot{m}_{\mathrm{FM^{TP}_{\textit{p}}}} + \dot{m}_{\mathrm{FM^{FP}_{\textit{p}}}},\nonumber\\
	\mathrm{where}\: p &\in \{\mathrm{S, M, L}\},\nonumber
	\label{equ:MPC_Loss}
\end{align}

where, $\dot{m}_{\mathrm{NFM_{Total}}}$ represents the total non-ferromagnetic mass flow outputted in a particular outlet of the magnetic sorter; constituted by the sum of the non-ferromagnetic mass flows of three size classes (small, medium, large). For each size class, the mass flow $\dot{m}_{\mathrm{NFM_{\textit{p}}}}$, where the variable $p$ stands for "size", is constituted by the correctly sorted mass flow $\dot{m}_{\mathrm{NFM^{TP}_{\textit{p}}}}$ and the incorrectly sorted mass flow $\dot{m}_{\mathrm{NFM^{FP}_{\textit{p}}}}$. Identical symbology is used for the FM material.

The loss function only directly references material compositions available at the magnetic sorter. However, in the backpropagation the ANT passes the gradients on through the conveyor belt to the sieving machine, thereby correctly treating dead times during material transport.

Instead of gradient descent we use RProp \citep{Riedmiller1992RPROPA,Igel2000ImprovingTR} because it features an adaptive step size that is independent of the loss gradient magnitude. Hence, the stability of the local node is independent of the loss function design.

\subsection{Evaluation} 

In order to validate our concept we designed two scenarios; one where the input material stream is constant and one in which it changes following a rectangular function. A detailed description of the input conditions is presented in Appendix \ref{Apdx:Virtual_facility}.

Figure \ref{fig:mass_flow_static}, shows selected mass flow fractions for the constant scenario. The inferred mass flows (lines) for all  machines match the ground truth virtual measurements (dots) within the predicted uncertainty, despite the ANT only being supplied total mass flows. The interval with growing mass flow uncertainties on the right end of the figures represents the ANT predictions at the end of the recording. For the conveyor belt in Figure \ref{fig:mass_flow_static} the dead time of 32~s between input and output is clearly visible as a shift by one time step.

After one real-time step transition and the introduction of new sensor information into the nodes, state inference for a single time step converged (below $\Delta_\mathrm{info}$) after a maximum of three information messages exchanged with its neighbors. Summing over all sensor data induced updates, up to 24 information messages are exchanged between two neighboring ANT nodes on the state of a single time step. Of course, these numbers are heavily dependent on the non-linearities present in used process models.

After 5 minutes the ANT machine parameter optimization is activated (green line). For the magnetic sorter Figure \ref{fig:mass_flow_static}(c), the magnet height approaches the optimum value of 11.2~cm (dashed line). The effect is clearly reflected in the correctly sorted FM mass flow (True Positive FM material) almost reaching the inputted mass flow. Due to the mass flow sensor data noise of approximately 18~$\%$, and the thereby induced uncertainty of the state estimates, a fluctuation in the magnet height remains even after convergence, featuring a mean value of 11.0~cm and a standard deviation of 0.35~cm.

Instead of reaching a loss minimum the sieving machine hits the parameter boundary at 21~rpm and therefore it does not show any more fluctuations. Here the effect on the mass flow is less obvious and on the order of the inferred uncertainty.

\begin{figure*}
	\centerline{\includegraphics[scale=0.4]{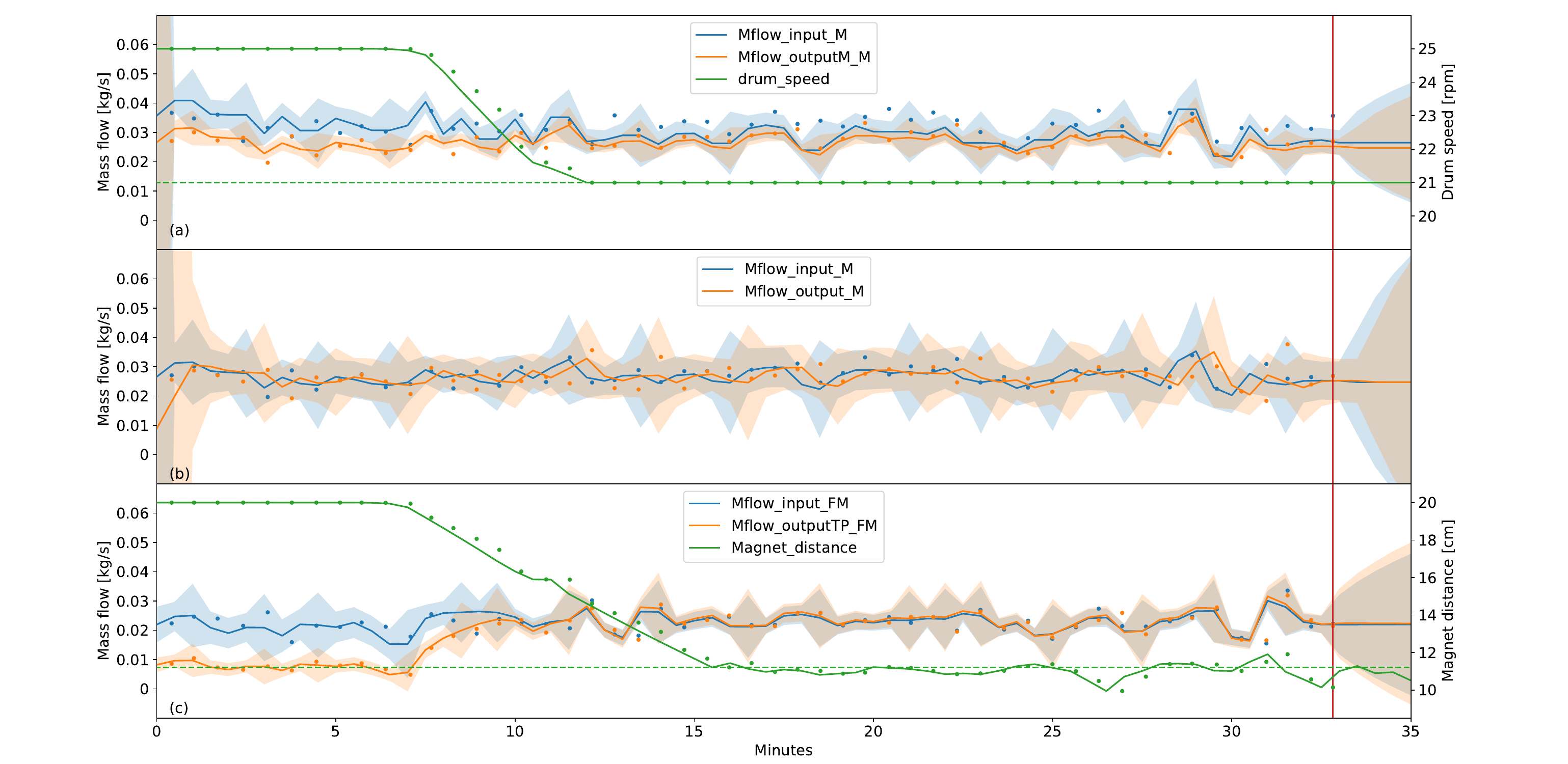}}
	\caption{Constant input mass flow scenario. Lines represent ANT  estimates, dots represent measurements from the virtual facility, the shaded error areas illustrate standard uncertainty of the ANT inference, the dashed lines indicate the calculated optimum value and the red line indicates data acquisition end (present). (a) Sieving machine, (b) conveyor belt, (c) Magnetic sorter.}
	\label{fig:mass_flow_static}                             
\end{figure*}
\begin{figure*}
	
	\centerline{\includegraphics[scale=0.4]{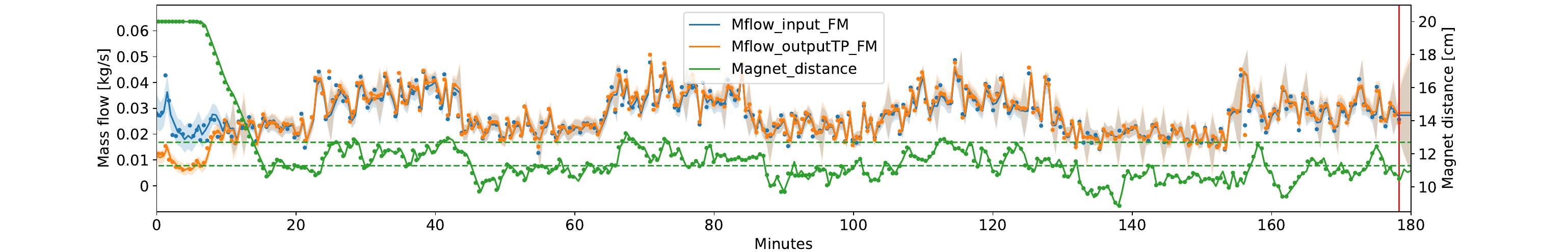}}
	\caption{Dynamic input mass flow scenario on the example of the magnetic sorter. Lines represent ANT  estimates, dots represent measurements from the virtual facility, the shaded error areas illustrate standard uncertainty of the ANT inference, the dashed lines indicate the calculated optimum value}
	\label{fig:mass_flow_dynamic}                             
\end{figure*}

Figure \ref{fig:mass_flow_dynamic} demonstrates the ANT with variable input mass flows on the example of the magnetic sorter. As shown by the inferred mass flows the ANT is able to resolve the steps in mass flow composition during inference. The two mass flow settings yield two distinct optima presented as dashed lines at 11.3~cm and 12.7~cm, calculated under the assumption of stationary mean mass flows and no model error. The ANT reproduces the expected jumps between the optima, but never fully converges to an optimum, which is again attributed to the noisy state estimate and possibly limited convergence rate. For the lower optimum interval the ANT yields a mean value of 10.8~cm and a standard deviation of 0.4~cm. For the intervals at the upper optimum the ANT achieves a mean value of 11.8~cm and a standard deviation of 0.6~cm. While, the strong remaining fluctuation due to state estimation noise barely allows for the distinction of the two optima, this still demonstrates that the ANT is capable to keep the system close to the optimum even in an noisy environment. 

\section{Discussion and Conclusion}
The ANT joins the concepts of MPC, sensor networks and backpropagation, to create a generalized framework for decentral process monitoring and optimization/control. This is achieved by formulating the statistical data fusion encountered in sensor networks or Kalman filtering in a differentiable manner, facilitating the decentral backpropagation of gradients and thus optimization of machine parameters or even fine-tuning local AI models.
The decentral nature guarantees data sovereignty of each node or actor, meaning no process model, internal state information or flow chart needs to be shared apart from each node knowing its neighbors. \par 
The ANT is applicable to any differentiable process chain, i.e. consisting of processes that can be modeled in a differentiable manner. However, the question if a process is differentiable is often a question of required temporal or quantitative resolution: if single articles need to be tracked which either do or do not have a certain property, the process can not be regarded as differentiable. If the expected fraction of articles with a certain property is sufficient to describe the process state, then the process can be modeled in a differentiable manner. This is also demonstrated by the plastic waste sorting example which essentially also processes discrete articles, but only the expected mass fractions are of interest.  \par 
Due to the generality of the ANT, it promises not only to facilitate process optimization, but analogously also retrain AI-based sensors or process models. Thus, in case of a drift in process behavior or material constituents this can avoid the recreation of expensive training datasets. While the implementation of continuous AI-model retraining through the ANT is beyond the scope of this work, it is still presented as one of the core motivations behind the ANT concept.\par 
Process monitoring and optimization capabilities are demonstrated on a virtual plastic waste sorting process, that allows for direct capture of ground truth process data in order to validate the ANT. The ANT, which is only provided total mass fluxes with approximately 18~$\%$ relative noise, was able to infer mass flow fractions reliably within estimated uncertainties. Process optimization converged close to the expected optima for constant and dynamic mass flow scenarios. The remaining noise of the state estimations causes a residual fluctuation in the optimization parameters. This effect can be interpreted as the result of noisy loss gradients or layer jacobians, which are typically not encountered in standard deep-learning, but might be treatable with tailored optimization methods.\par

A comparison to a standard MPC approach is omitted, as the same cost function is optimized, and a centralized MPC setting can be expected to reach the same optimum significantly faster (neglecting data sovereignty). The present demonstration is meant as a proof of principle. In the future, optimized ANT implementations are to be bench-marked.\par 
While meant to provide the possibility of continuous model retraining, the ANT is model agnostic, meaning anything from white-box physical models to black-box neural networks can be employed. As such the ANT can also be viewed as a universal model connector. In the special case of connecting black and white box models it can as well be seen as a special form of operator based learning \citep{maier2022known,maier2019learning}. Likewise similar problems are treated on the boundary between mathematical optimization and control theory \citep{2021arXiv210311329H}. \par 
As explained in the introduction, we aim to be able to design ANT nodes independently of one another or the loss node, without impacting function or stability. Thereby we want to alleviate a restriction often imposed in distributed/decentral MPC. In the simple most form of the ANT, using gradient descent with a fixed learning rate, this goal is certainly not fulfilled, as scaling the loss function, or a neighbor node exhibiting increased gradient gains, would easily impact stability. Further, in practical applications each node must be safe by design, with defined parameter ranges and maximum parameter step sizes. This is different from deep-learning topics where typically no such information is available. Using this prior information and more advanced optimizer algorithms with adaptive learning rates can bring us closer to the goal of fully independent node design. Therefore, the virtual demonstration used the RProp algorithm \citep{Igel2000ImprovingTR} for its adaptive step size which is independent of gradient magnitude, and thus choice and scaling of the loss function. Further, boundaries as well as minimum and maximum step sizes can easily be applied using RProp. In summary, the ANT transfers the necessity to ensure stability from the MPC loop design to the design of every single node. The actual choice of optimizer is thus highly dependent on the process controlled by the node, and possibly its safety requirements.\par 
Just as many insights on stability can be drawn from control theory \citep{2021arXiv210311329H} or deep-learning literature, many other advanced methods can be included that are already available in the control, sensor network or deep-learning community. Another example could be the generalization of the method beyond MVN probability models through deep-learning methods \citep{Horger2018} or smart transformations \citep{Hanebeck2003}.  The presented method only seeks to represent a minimum skeleton to join these fields. The results are regarded as very promising, making the next steps further development of the concept and design of a physical demonstrator.

\section{Funding}                              
Funded by German Ministry of Education and Research (BMBF) under grant number 033KI201.

\bibliographystyle{elsarticle-harv}
\bibliography{lit}

\appendix
\section{Implicit MAP estimate derivative}   
\label{app:implderiv}
According to the implicit function theorem, function $g$ exists if $\mathbf{k}(\mathbf{x}^*,\boldsymbol\beta)$ is continuously differentiable and the Jacobian of $\mathbf{k}$ with respect to $\mathbf{x}^*$ is invertible. The differentiability is assumed to be fullfilled because we require the models to be continuously differentiable and the linear measurement equations as well as the sum of squares are continuously differentiable. The invertibility of $\mathcal{J}_{\mathbf{k},\mathbf{x}^*}$ is provided if the posterior is non-degenerate and thus needs to be checked anyway.\par
Provided these conditions, the sought Jacobian of the MAP state vector can be expressed,
\begin{equation}
	\mathcal{J}_{\mathbf{g},\boldsymbol\beta} = - \mathcal{J}_{\mathbf{k}, \mathbf{x}^*}^{-1} \mathcal{J}_{\mathbf{k}, \boldsymbol\beta}.
\end{equation}
To determine the required Jacobians, $\mathcal{J}_{\mathbf{k}, \mathbf{x}^*}= \frac{\partial \mathbf{k}}{\partial \mathbf{x}^*}$ and $\mathcal{J}_{\mathbf{k}, \boldsymbol\beta}= \frac{\partial \mathbf{k}}{\partial \boldsymbol\beta}$, we express $\mathbf{k}$ explicitly,
\begin{equation}
	\mathbf{k}(\mathbf{x}^*,\boldsymbol\beta)=\frac{\partial\chi^2}{\partial \mathbf{x}^*}(\mathbf{x}^*,\boldsymbol\beta) =2\left[ \frac{\partial \mathbf{f}}{\partial \mathbf{x}^*}(\mathbf{x}^*,\boldsymbol\beta) \right]^T \mathbf{f}(\mathbf{x}^*,\boldsymbol\beta).
\end{equation}
In this form the Jacobians become,
\begin{align}
	\mathcal{J}_{\mathbf{k}, \mathbf{x}^*}= \frac{\partial \mathbf{k}}{\partial \mathbf{x}^*}= 2\left[\frac{\partial}{\partial\mathbf{x}^*}\left(\frac{\partial\mathbf{f}}{\partial\mathbf{x}^*}\right)^T\right] \mathbf{f}+ 2\left(\frac{\partial\mathbf{f}}{\partial \mathbf{x}^*}\right)^T\frac{\partial\mathbf{f}}{\partial \mathbf{x}^*},\\
	\mathcal{J}_{\mathbf{k}, \boldsymbol\beta}= \frac{\partial \mathbf{k}}{\partial \boldsymbol\beta}= 2\left[\frac{\partial}{\partial\boldsymbol\beta}\left(\frac{\partial\mathbf{f}}{\partial\mathbf{x}^*}\right)^T\right] \mathbf{f}+ 2\left(\frac{\partial\mathbf{f}}{\partial \mathbf{x}^*}\right)^T\frac{\partial\mathbf{f}}{\partial \boldsymbol\beta}.
\end{align}
To make the calculation tractable, second order terms are omitted,
\begin{align}
	\mathcal{J}_{\mathbf{k}, \mathbf{x}^*}\approx  2\left(\frac{\partial\mathbf{f}}{\partial \mathbf{x}^*}\right)^T\frac{\partial\mathbf{f}}{\partial \mathbf{x}^*},\\
	\mathcal{J}_{\mathbf{k}, \boldsymbol\beta}\approx  2\left(\frac{\partial\mathbf{f}}{\partial \mathbf{x}^*}\right)^T\frac{\partial\mathbf{f}}{\partial \boldsymbol\beta},
\end{align}
yielding the implicit Jacobian,
\begin{align}
	\mathcal{J}_{\mathbf{g},\boldsymbol\beta} \approx& -\left[ \left(\frac{\partial\mathbf{f}}{\partial \mathbf{x}^*}\right)^T\frac{\partial\mathbf{f}}{\partial \mathbf{x}^*} \right]^{-1} \left(\frac{\partial\mathbf{f}}{\partial \mathbf{x}^*}\right)^T\frac{\partial\mathbf{f}}{\partial \boldsymbol\beta}=\nonumber\\
	=& \left[ \left(\mathbf{L}\mathcal{J}_{\mathbf{F},\mathbf{x}^*}\right)^T\mathbf{L}\mathcal{J}_{\mathbf{F},\mathbf{x}^*} \right]^{-1} \left(\mathbf{L}\mathcal{J}_{\mathbf{F},\mathbf{x}^*}\right)^T\mathbf{L}
\end{align}
\section{ANT algorithm from node perspective}
\label{app:ant_alg}
The following listing describes the behavior of an ANT node in response to external messages.
\begin{enumerate}
	\item Initialize in Information period and unfrozen state.
	\item Ensure a sufficient finite horizon, by appending necessary time steps.
	\item Include newly available sensor data into the corresponding time steps.
	\item Receive ANT messages:
	\begin{enumerate}
		\item Process all information messages, by integrating them into the corresponding time step. Already existing information from the same neighbor for the same time step is overwritten.
		\item If in Information period, resolve updated time steps
		\item Process control messages:
		\begin{enumerate}
			\item If switching to information period: apply gradient descent update, unfreeze, resolve all necessary steps and set Information period.
			\item If switching to backpropagation period: freeze the state, zero gradient, set Backpropagation period.
		\end{enumerate} 
		\item Process gradient messages. Backpropagate them through time and to the corresponding gradient storages of each measurement and model, and agglomerate there.
	\end{enumerate}
	
	\item Generate messages:
	\begin{enumerate}
		\item If in Information period: for all neighbors and all time steps:
		\begin{enumerate}
			\item If no information has been sent to this neighbor and time step or if the KL divergence between the sent information and the newly available estimate is larger than $\Delta_\mathrm{info}$ (e.g. $10^{-4}$~bits), send an (updated) information message.
		\end{enumerate} 
		\item If in Backpropagation period: for all informations from neighbors in all time steps:
		\begin{enumerate}
			\item If any of the elements of $\Delta_\mathrm{grad}$ times the largest gradient sent for this time step is smaller than the gradient currently in gradient storage, send the gradient form gradient storage to the corresponding neighbor. 
		\end{enumerate}
	\end{enumerate}
	\item Set the machine parameters according to the parameter stored in the current time step.
	\item Go to (2)
\end{enumerate}

\section{Virtual facility}\label{Apdx:Virtual_facility}

\subsection{Simulated Materials}

The represented sorting processes (size and magnetic sorting) require a material set representative enough so that many scenarios can be emulated, including missing materials for a specific class. Towards that, we introduced three sizes of objects: small, medium, and large; which are divided into ferromagnetic, and non-ferromagnetic. The full description of the materials used can be found in Table \ref{tab:materials}.

\begin{figure}[h]
	\centerline{\includegraphics[scale=0.2]{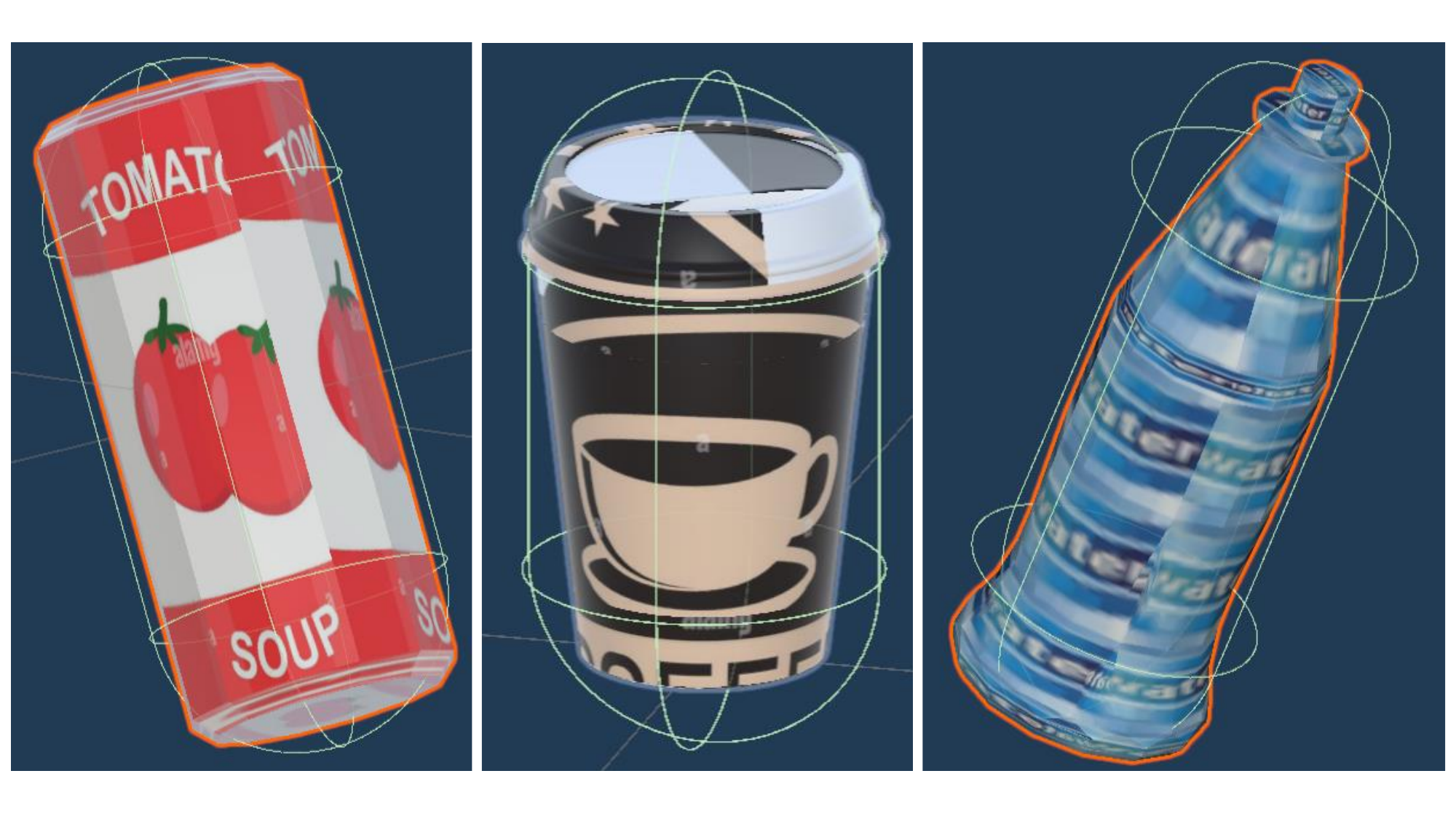}}
	\caption{Cola can (Left), coffee cup (Middle), plastic bottle (Right)}
	\label{fig:MS_materials}
\end{figure}

\begin{table*}[h]
	\caption{Recycling materials properties.}
	\label{tab:materials}
	\begin{center}
		\begin{tabular}{l|ccccccc}
			\multicolumn{1}{c|}{}&\multicolumn{1}{c|}{} & \multicolumn{2}{c}{\textbf{Mean mass flow [kg/s]}}&\multicolumn{1}{c|}{}&\multicolumn{1}{c|}{}&\multicolumn{1}{c}{}\\ 
			\hline\hline
			\multicolumn{1}{c|}{\textbf{Name}}&\multicolumn{1}{c|}{\textbf{Mass [kg]}} & \multicolumn{1}{c|}{\textbf{Static Scenario}} & \multicolumn{2}{c|}{\textbf{Dynamic scenario}} & \multicolumn{1}{c|}{\textbf{Ferromagnetic class}} &\textbf{Size class}\\\cline{1-7}
			
			\multicolumn{1}{c|}{Paper roll}&\multicolumn{1}{c|}{0.0090} & \multicolumn{1}{c|}{0.0069} & \multicolumn{1}{c|}{0.0520} &\multicolumn{1}{c|}{0.0199} & \multicolumn{1}{c|}{NFM} &{Large}\\
			\multicolumn{1}{c|}{Plastic bottle}&\multicolumn{1}{c|}{0.0180} & \multicolumn{1}{c|}{0.0275} & \multicolumn{1}{c|}{0.0012} &\multicolumn{1}{c|}{0.0348} & \multicolumn{1}{c|}{NFM} &{Large}\\
			\multicolumn{1}{c|}{Coffee cup}&\multicolumn{1}{c|}{0.0093} & \multicolumn{1}{c|}{0.0068} & \multicolumn{1}{c|}{0.0003} &\multicolumn{1}{c|}{0.0087} & \multicolumn{1}{c|}{NFM} &{Medium}\\
			\multicolumn{1}{c|}{Paper ball}&\multicolumn{1}{c|}{0.0035} & \multicolumn{1}{c|}{0.0026} & \multicolumn{1}{c|}{0.0001} &\multicolumn{1}{c|}{0.0033} & \multicolumn{1}{c|}{NFM} &{Medium}\\
			\multicolumn{1}{c|}{FM can}&\multicolumn{1}{c|}{0.0149} & \multicolumn{1}{c|}{0.0226} & \multicolumn{1}{c|}{0.0356} &\multicolumn{1}{c|}{0.0226} & \multicolumn{1}{c|}{FM} &{Medium}\\
			\multicolumn{1}{c|}{FM cap}&\multicolumn{1}{c|}{0.0007} & \multicolumn{1}{c|}{0.0011} & \multicolumn{1}{c|}{0.0001} &\multicolumn{1}{c|}{0.0013} & \multicolumn{1}{c|}{FM} &{Small}\\
			\multicolumn{1}{c|}{NFM cap}&\multicolumn{1}{c|}{0.0005} & \multicolumn{1}{c|}{0.0007} & \multicolumn{1}{c|}{0.0012} &\multicolumn{1}{c|}{0.0008} & \multicolumn{1}{c|}{NFM} &{Small}\\
			
		\end{tabular}
	\end{center}
\end{table*}
While the simulation of a real world process is not the goal of the virtual facility, this set of materials already allows us to reach a realistic level of complexity in terms of process modeling, with problems like: materials falling out of conveyor belts, getting stuck between machine elements or as mentioned before, the inclusion of impurity on the sorting outputs caused by the normal behavior of a machine whose parameters have not been optimized. A sample of the set of objects used in Unity is depicted in Figure \ref{fig:MS_materials}.

\subsection{Simulated Machines}

\subsubsection{Sieving machine}

Objects are introduced in the simulation environment inside the black duct displayed in the upper left corner of Figure \ref{fig:machines_diagram}. The number of articles instantiated per class and time step is sampled from a Poisson distribution.

After being introduced, these objects are counted in the "Input Sensor" during fixed time intervals of 30~s. Class specific mass flows are calculated from article count, mass and time interval.

The sieving machine has three filtered sizes: small, medium and large. Accordingly, the cylindrical drum sieve has 3 sections, each with a different pattern of holes: the first one (S) that will separate objects of sizes up to 2.5cm; the second (M), which will allow the transit of objects smaller than 12.5cm and the last one (L) with 25cm diameter holes. The drum has an inclination of 12.5 degrees which facilitates the material flow inside of it. Additionally, the drum rotates with an adjustable speed, which is this machine's optimization parameter.

The collector is placed under the drum, to collect the filtered materials and direct them to the specific conveyor belt, which will transport them to the corresponding next machine in the process.

\subsubsection{Long conveyor belt}
In order to require the ANT to backpropagate gradients through time, the conveyor belts are designed to introduce a dead time of 32~s, which is larger than the ANT time step size used. Material mixing on the conveyor belt is negligible.

\subsubsection{Magnetic sorter}
The magnetic sorter separates ferromagnetic and non-ferromagnetic materials as shown in Figure \ref{fig:machines_diagram} For simulating the effects of a magnetic field, we created a force field that affects all ferromagnetic objects (FM caps and cans) that reach the area of influence of the field and pushes them towards the uppermost, magnetic conveyor belt. The magnetic conveyor belt drags them forward and releases them on the upper outlet conveyor belt where the FM mass flow is determined by the output FM sensor. 

The strength of the force field is inversely proportional to the magnet-object distance squared. The absolute height is the optimization parameter as for the real world magnetic sorter.


On the other hand, NFM materials, pass through the input sensor and fall directly to the lower conveyor belt where they are transported towards the respective sensor, and their mass flow is determined.

As it is in reality, this sorting process is not perfect, so impurity is present in both streams of materials. In the upper outlet, when the magnet distance is low, the acceleration given to the metallic objects by the force field is very high, causing some NFM objects to be dragged to the upper output as well. if the magnet too high up, the force applied to the FM objects is insufficient to carry them to the upper belt, resulting in an increased FM content in the lower NFM output. Between these two points there is an optimum distance where the majority of the FM objects are transported to the FM outlet without considerable impurity, and therefore most NFM objects are transported to the NFM outlet. The exact position of this optimum is depending on the weigting applied in the loss function.

\section{Process models}\label{Apdx:process_models}
As long as the models are differentiable, the ANT is capable of embedding different kinds of models, in terms of how much knowledge about the system or training data is available. In order to demonstrate this capability we decided to build our system including: black box, gray box and white box models.

\subsection{Black box model}
For the magnetic sorting machine a black box model was implemented. Here a feed forward neural network was used, with its characteristic input (Input mass flow, plus magnet distance), two hidden, and output layer (true and false positives for FM and NFM outputs respectively). This network was trained on a data set that was generated in the previously mentioned virtual facility, which includes all the features required by the model (Input mass flows FM/NFM, magnet distance, true and false positive FM, true and false positive NFM). The model is therefore ignorant to size classes.

\begin{figure*}[h]
	\centerline{\includegraphics[scale=0.35]{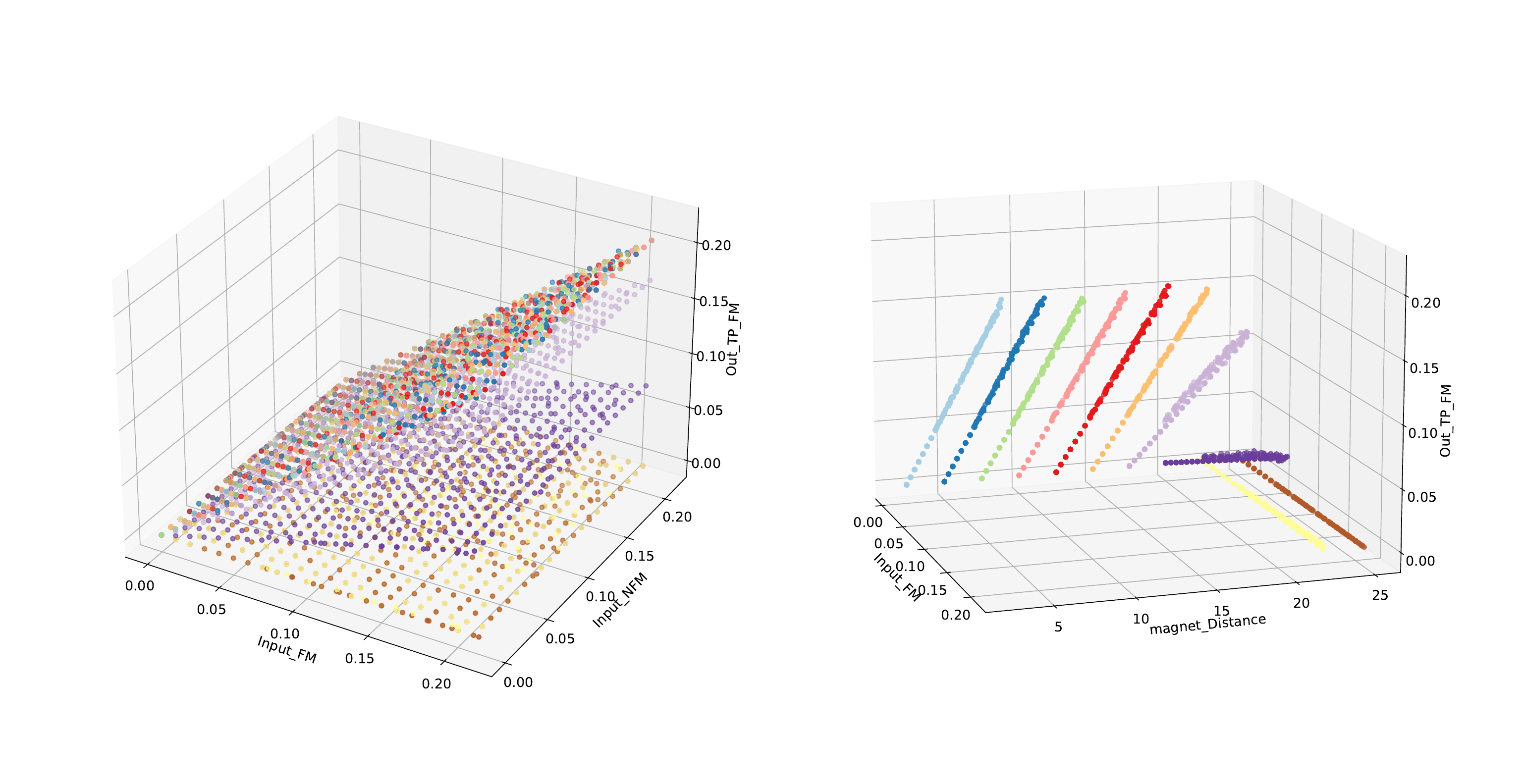}}
	\caption{Data set for magnetic sorter model, data distribution depending on the input (left), data distribution depending on the magnet distance (right)}
	\label{fig:MS_dataset}
\end{figure*}

Figure \ref{fig:MS_dataset} depicts the distribution of the data used to train the black box model, where the colors represent the different magnet distances in which data was collected as the input mass flows variate within a predefined range. On the right side of the mentioned figure, the FM outputs dependency on the magnet distance is illustrated. The covariance of model uncertainty (see $\boldsymbol\epsilon_\mathrm{proc}$ in Equation \eqref{equ:impl_proc_model}) is estimated from the residual of the training.

To connect the size class based material description of the siever to the ferromegnetism based description used by the magnetic sorter additional relations are introduced: the assumption of mass conservation for each size class, as well as assumptions of statistical independence between size class and ferromagnetism. For a size class $p$, mass conservation is modeled by,
\begin{align}
	\dot{m}_{\mathrm{in^{NFM}_{\textit{p}}}}
	&=\dot{m}_{\mathrm{out^{NFM}_{NFM_\textit{p}}}} + \dot{m}_{\mathrm{out^{NFM}_{FM_\textit{p}}}}\label{eq:mass_conserv1} +\epsilon_1,\\
	\dot{m}_{\mathrm{in^{FM}_{\textit{p}}}}
	&=\dot{m}_{\mathrm{out^{FM}_{NFM_\textit{p}}}} + \dot{m}_{\mathrm{out^{FM}_{FM_\textit{p}}}}\label{eq:mass_conserv2}+\epsilon_2,\\
	\mathrm{where}\: p &\in \{\mathrm{S, M, L}\}, \nonumber
\end{align}
where the inputted mass flow for a material type NFM/FM is equal to the output mass flow of these material type in both outlets of the magnetic sorter. The independent, unbiased model uncertainties $\epsilon_1$, $\epsilon_2$ are chosen to have a variance of $10^{-4}$~$\mathrm{kg}^2/\mathrm{s}^2$.
Statistical independence at the input is enforced via,
\begin{align}
	\dot{m}_{\mathrm{in^{NFM}_{S}}}& ( \dot{m}_{\mathrm{in^{NFM}_{M}}}+\dot{m}_{\mathrm{in^{FM}_{M}}})              =\nonumber\\
	&\dot{m}_{\mathrm{in^{NFM}_{M}}}   (\dot{m}_{\mathrm{in^{NFM}_{S}}}+\dot{m}_{\mathrm{in^{FM}_{S}}}) +\epsilon_3   \label{eq:mass_S_M}\\
	\dot{m}_{\mathrm{in^{NFM}_{M}}}  &    ( \dot{m}_{\mathrm{in^{NFM}_{L}}}+\dot{m}_{\mathrm{in^{NFM}_{L}}})            =\nonumber\\
	&\dot{m}_{\mathrm{in^{NFM}_{L}}}   (\dot{m}_{\mathrm{in^{NFM}_{M}}}+\dot{m}_{\mathrm{in^{FM}_{M}}})   +\epsilon_4        \label{eq:mass_M_L}
\end{align}
where, $\dot{m}_{\mathrm{in^{NFM}_{S}}}$ represents the inputted mass flow of non-ferromagnetic material of small size. 
Analogously the outputs are constrained to be independent by,

\begin{align}
	\dot{m}_{\mathrm{out^{NFM}_{NFM_\textit{p}}}} \dot{m}_{\mathrm{out^{NFM}_{FP_{Total}}}}             
	=\dot{m}_{\mathrm{out^{NFM}_{FM_\textit{p}}}}   \dot{m}_{\mathrm{out^{NFM}_{TP_{Total}}}}      +\epsilon_{p,1},       \label{eq:mass_NFM}\\
	\textnormal{with }\dot{m}_{\mathrm{out^{NFM}_{TP_{Total}}}}
	=\dot{m}_{\mathrm{out^{NFM}_{NFM_S}}}+\dot{m}_{\mathrm{out^{NFM}_{NFM_M}}}+\dot{m}_{\mathrm{out^{NFM}_{NFM_L}}},\nonumber\\
	\dot{m}_{\mathrm{out^{NFM}_{FP_{Total}}}}
	=\dot{m}_{\mathrm{out^{NFM}_{FM_S}}}+\dot{m}_{\mathrm{out^{NFM}_{FM_M}}}+\dot{m}_{\mathrm{out^{NFM}_{FM_L}}},\nonumber\\
	\textnormal{and}\nonumber\\
	\dot{m}_{\mathrm{out^{FM}_{NFM_\textit{p}}}}       \dot{m}_{\mathrm{out^{FM}_{TP_{Total}}}}          
	=\dot{m}_{\mathrm{out^{FM}_{FM_\textit{p}}}}      \dot{m}_{\mathrm{out^{FM}_{FP_{Total}}}}     +\epsilon_{p,2},   \label{eq:mass_FM}\\
	\textnormal{with }\dot{m}_{\mathrm{out^{FM}_{TP_{Total}}}}
	=\dot{m}_{\mathrm{out^{FM}_{FM_S}}}+\dot{m}_{\mathrm{out^{FM}_{FM_M}}}+\dot{m}_{\mathrm{out^{FM}_{FM_L}}}\nonumber\\
	\dot{m}_{\mathrm{out^{FM}_{FP_{Total}}}}
	=\dot{m}_{\mathrm{out^{FM}_{NFM_S}}}+\dot{m}_{\mathrm{out^{FM}_{NFM_M}}}+\dot{m}_{\mathrm{out^{FM}_{NFM_L}}}\nonumber\\
	\mathrm{where}\: p \in \{\mathrm{S, M, L}\}. \nonumber
\end{align}
The variance of model uncertainty, $\epsilon_i$, $\epsilon_{p,i}$, for the independency equations is set to $10^{-8}$~$\mathrm{kg}^4/\mathrm{s}^4$.

In absence of a physical motivation for a prediction model (no dynamic dependence), a simple low pass model was used as prediction model, only weakly restraining changes in mass flows between time steps with a standard deviation of $10^{-2}$~$\mathrm{kg}/\mathrm{s}$ for the inputs and $1$~$\mathrm{kg}/\mathrm{s}$

\subsection{White box model}

As mentioned in the result Section \ref{sec:Results}, conveyor belts were modeled as white box models. To reproduce the expected time dead time, $\delta_t=32$~s, the state vectors contain a history of input mass flows during prior time steps. The prediction model states that items in the history of the current time steps's state are shifted by one time step in the next time steps' state vector similar to a shift register,

\begin{align}
	(\dot{m}_{t})_h = (\dot{m}_{t+1})_{h+1}+ \epsilon_h,
\end{align}
where $\dot{m}_{t}$ is the mass flow history in the current state vector, $\dot{m}_{t+1}$ is the mass flow history in the next time steps state vector, $h$ is the index in the mass flow history, and $\epsilon_h$ is the model uncertainty with a standard deviation of $10^{-4}$~$\mathrm{kg}/\mathrm{s}$. This shift register like operation is implemented for all three size classes received by the siever. The output is the linear interpolation of the mass flow history at the point corresponding to the dead time.

\subsection{Gray box model}\label{sec-apx:Gray_box_model}
The sieving machine as mentioned before was modeled as a gray box model, which consists of two parts: mixing and dead time inside the sieving drum, and the rotational-speed dependence of the output fractions.

\begin{figure*}[h]
	\centerline{\includegraphics[scale=0.35]{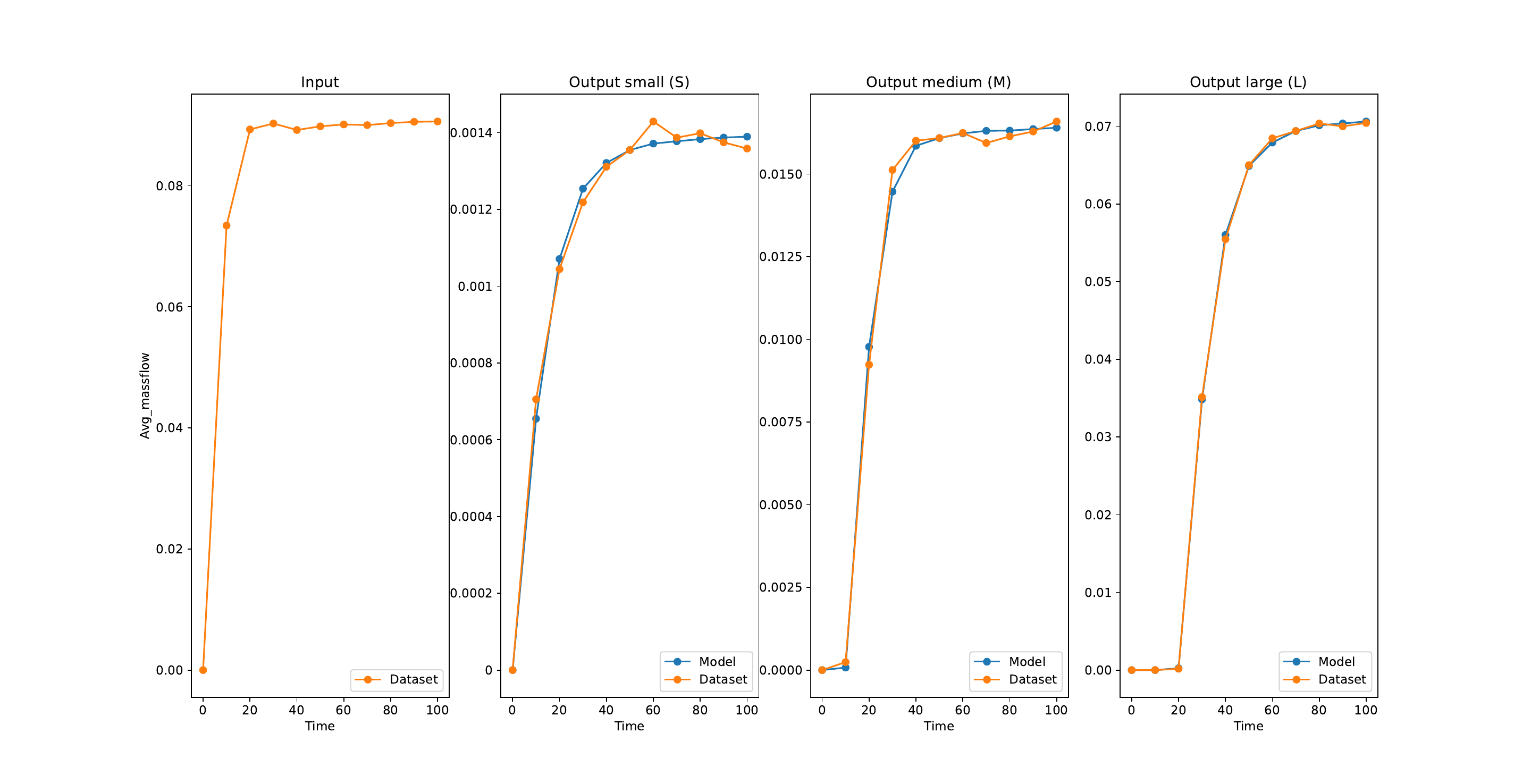}}
	\caption{Sieving machine, mixing behavior model (Orange), after an input function (left) convoluted with an impulse function to obtain the respective outputs (Blue):Small, Medium, and Large.}
	\label{fig:Siever_Mix}
\end{figure*}

Mixing and dead time are modeled analogously to the conveyor belt model by implementing a history of input mass flows in the state vector. While linear interpolation is sufficient to model the dead time, the mixing needs a more elaborate model. Due to their differentiability and simple implementation we rely on methods from reactor network modeling \citep{Liotta2014} in which dead time and mixing are treated at the same time by convolution of the history with a residence time distribution similar in nature to an impulse response function. The residence time distribution is modeled as an exponential decay with mixing time constant $\tau$ (well-stirred reactor \citep{Liotta2014}), shifted to the right by the dead time, $\delta_t$ (plug flow reactor\citep{Liotta2014}). 
The required parameters required to model the dynamic behavior of each sieving drum output , $\delta_t$ and $\tau$, were determined from a model fit to measured step responses of the siever as illustrated in Figure \ref{fig:Siever_Mix}.

\begin{figure*}[h]
	\centerline{\includegraphics[scale=0.35]{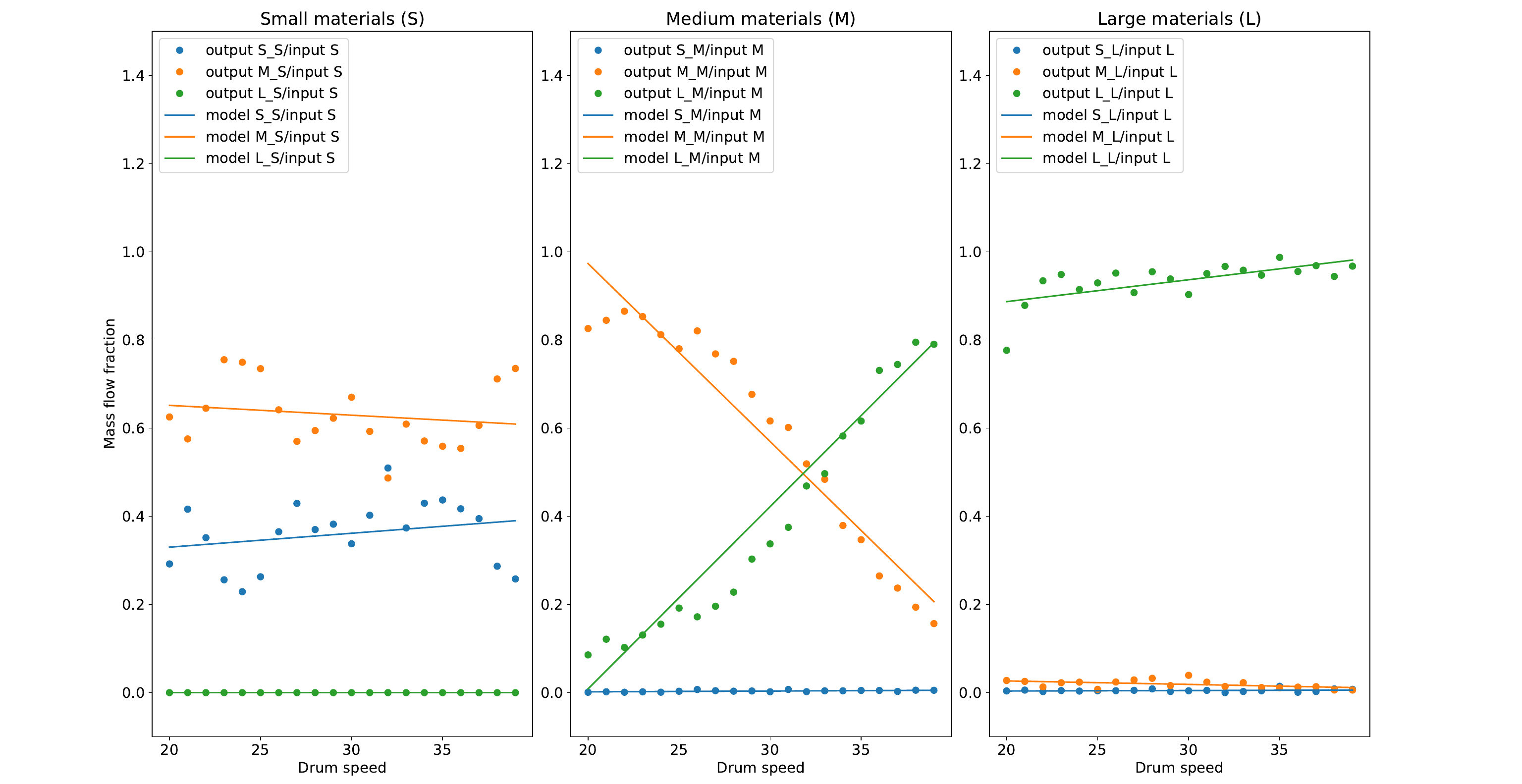}}
	\caption{Sieving machine, speed dependence of each size of material. Mass flow fractions on three outlets for small materials (left), medium materials (Center), and Large materials (Right)}
	\label{fig:Siever_speed}
\end{figure*}

The size class mass flow fractions arriving at each output are dependent on the rotational speed of the sieving drum. Accordingly, measurements of the mass ratios at different rotation speeds were acquired in the virtual facility, and plotted in Figure \ref{fig:Siever_speed}.

In figure \ref{fig:Siever_speed} contains three plots each referring to one size class of material. Each plot show the fraction of the size classes' input material, which through each corresponding output. For example the orange data points in the left plot describe the small size mass flow leaving through output M divided by the small size mass flow input into the siever. The noisy data is modeled by linear relationships with respect to the drum speed as also shown in the figure. The middle size class material shows the largest dependence on rotational speed, which is also why it was selected as a target in the loss function.

\end{document}